\documentclass[journal]{IEEEtran}
\usepackage{mathrsfs}
\usepackage{amsmath}
\usepackage{multirow}
\usepackage{booktabs}
\usepackage{pifont}       
\usepackage{bbding}       
\usepackage{tikz}
\usetikzlibrary{shapes,arrows,calc,positioning,fit,backgrounds}
\usepackage{fontawesome}
\usepackage{xcolor}
\usepackage{tikz}
\definecolor{agentsecred}{RGB}{255, 102, 102} 
\definecolor{agentblue}{RGB}{102, 178, 255}  
\definecolor{pblue}{RGB}{173, 216, 230}      
\definecolor{bblue}{RGB}{135, 206, 250}      
\definecolor{actblue}{RGB}{100, 149, 237}    
\definecolor{envgreen}{RGB}{144, 238, 144}   
\usepackage[edges]{forest}
\definecolor{maincolor}{HTML}{4F81BD}
\definecolor{subcolor1}{HTML}{C0504D}
\definecolor{subcolor2}{HTML}{9BBB59}
\definecolor{subcolor3}{HTML}{8064A2}
\definecolor{subcolor4}{HTML}{4BACC6}
\definecolor{subcolor5}{HTML}{F79646}

\ifCLASSINFOpdf
\else
\fi
\begin{document}
%
\title{Loki’s Dance of Illusions: A Comprehensive Survey of Hallucination in Large Language Models}
%
%
%

\author{Chaozhuo Li, Pengbo Wang, Chenxu Wang, Litian Zhang, Zheng Liu, Qiwei Ye, Yuanbo Xu, Feiran Huang, Xi Zhang, Philip S. Yu \IEEEmembership{Fellow, IEEE}
\IEEEcompsocitemizethanks{
\IEEEcompsocthanksitem Chaozhuo Li, Pengbo Wang, Chuanshi and Xi Zhang are with Beijing University of Posts and Telecommunications, Beijing, China (e-mail: lichaozhuo@bupt.edu.cn). Chenxu Wang is with Shihezi University, Shihezi, China. Litian Zhang is with Beihang University, Beijing, China. Zheng Liu and Qiwei Ye are with Beijing Academy of Artificial Intelligence, Beijing, China. Yuanbo Xu is with Jilin University, Jilin, China. Senzhang Wang is with Central South University, Hunan, China. Feiran Huang is with Jinan University, Guangdong, China. 
}
}

%
%

\markboth{ }%
{Shell \MakeLowercase{\textit{et al.}}: Bare Demo of IEEEtran.cls for IEEE Journals}
%



\maketitle

\begin{abstract}
Edgar Allan Poe noted, ``Truth often lurks in the shadow of error,'' highlighting the deep complexity intrinsic to the interplay between truth and falsehood, notably under conditions of cognitive and informational asymmetry. This dynamic is strikingly evident in large language models (LLMs). Despite their impressive linguistic generation capabilities, LLMs sometimes produce information that appears factually accurate but is, in reality, fabricated, an issue often referred to as 'hallucinations'. The prevalence of these hallucinations can mislead users, affecting their judgments and decisions. In sectors such as finance, law, and healthcare, such misinformation risks causing substantial economic losses, legal disputes, and health risks, with wide-ranging consequences.In our research, we have methodically categorized, analyzed the causes, detection methods, and solutions related to LLM hallucinations. Our efforts have particularly focused on understanding the roots of hallucinations and evaluating the efficacy of current strategies in revealing the underlying logic, thereby paving the way for the development of innovative and potent approaches. By examining why certain measures are effective against hallucinations, our study aims to foster a comprehensive approach to tackling this issue within the domain of LLMs.

\end{abstract}

\begin{IEEEkeywords}
LLM Hallucination Survey
\end{IEEEkeywords}

%
\IEEEpeerreviewmaketitle

\section{Introduction}
%
%
%
%

\IEEEPARstart{T}{he} advent of large language models (LLMs) has significantly advanced the field of natural language processing (NLP), enabling a broad spectrum of applications that extend beyond traditional text understanding and generation \cite{min2023recent}.
LLMs, characterized by their extensive parameterization and pretraining on massive corpora, have demonstrated remarkable proficiency in capturing complex linguistic patterns, contextual dependencies, and even certain forms of reasoning \cite{openai2023gpt4,guo2025deepseek}. 
The unprecedented scale and adaptability of LLMs indicate a paradigm shift in AI research, raising fundamental questions about the extent to which LLMs can generalize knowledge, demonstrate emergent reasoning capabilities, and bridge the gap between narrow task-specific intelligence and more generalized, human-like cognitive abilities \cite{tonmoy2024comprehensive}. 

\begin{figure}[t]
  \centering
  \includegraphics[width=0.95\columnwidth]{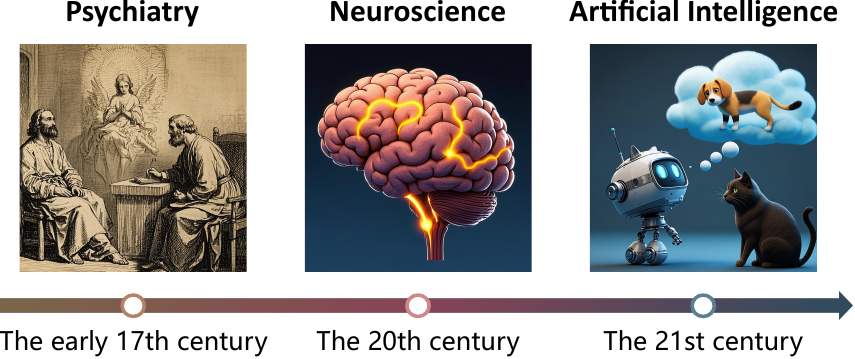} 
  \caption{\centering The evolutionary history of the term ``Hallucination''.} 
  \label{fig:intro1}
\end{figure}

Despite the remarkable success of LLMs, they remain burdened by a critical challenge—the persistent issue of \textit{hallucination}, akin to the proverbial \textit{Sword of Damocles} hanging over their reliability. As illustrated in Fig. \ref{fig:intro1}, the term ``hallucination'' originates from the Latin \textit{hallucinari}, meaning \textit{to wander in the mind} \cite{etymonline_hallucination}. 
In the 18th and 19th centuries, psychiatrists such as Jean-Étienne Dominique Esquirol refined its medical definition as ``false perceptions occurring without any external stimulus'' \cite{esquirol1838hallucinations}.  
By the 20th century, advancements in neuroscience linked hallucinations to mental disorders, including schizophrenia and Parkinson’s disease \cite{schizophrenia_parkinsons_hallucinations}. 


Recently, the term ``hallucination'' has gained prominence in artificial intelligence research \cite{koehn2017six, maynez2020faithfulness}. 
However, defining ``hallucination'' precisely in the context of LLMs remains challenging due to its diverse manifestations and the subjective nature of human interpretation. 
Previous studies generally first categorize hallucinations into subtypes and then  propose definitions within narrower conceptual frameworks. 
 Ji et al. \cite{ji2023survey} classify hallucinations into Intrinsic Hallucinations, where the output contradicts the source content, and Extrinsic Hallucinations, where the output cannot be verified against the source. 
 Banerjee et al. \cite{banerjee2024llms} propose a more detailed taxonomy, identifying Input-conflicting Hallucinations (deviations from user-provided input), Context-conflicting Hallucinations (contradictions with the model’s prior output), and Fact-conflicting Hallucinations (inaccuracies against established world knowledge).  
Our survey synthesizes foundational insights from contemporary literature \cite{ji2023survey, banerjee2024llms} and proposes a working definition of hallucinations in LLMs, drawing conceptual parallels with established medical phenomenology \cite{esquirol1838hallucinations}.
We formally define hallucinations as: \textit{Hallucinations refer to instances where LLMs produce seemingly plausible but factually inaccurate or contextually disconnected responses}. 
This definition emphasizes three important criteria: (1) surface-level plausibility, (2) factual or logical incongruence, and (3) failure to maintain contextual fidelity—dimensions that enable systematic analysis of hallucinatory patterns.

\begin{table*}[!ht]
\centering
\caption{A Comparative Analysis of Representative Surveys on Hallucinations in LLMs}
\label{tab:all_works1}
\setlength\tabcolsep{6.8pt}
\renewcommand{\arraystretch}{1.2} 
\resizebox{\linewidth}{!}{
\begin{tabular}{lcccccccc}
\toprule
\multirow{2}[2]{*}{\textbf{Objectives}} & \multicolumn{8}{c}{\textbf{Surveys on  Hallucinations}} \\ 
\cmidrule(lr){2-9}
& \textbf{Ours} & Tonmo et al. \cite{tonmoy2024comprehensive} & Rawte et al. \cite{rawte2023survey} & Zhang et al. \cite{zhang2023siren} & Huang et al. \cite{huang2023survey} & Bai et al. \cite{bai2024hallucination} & Jiang et al. \cite{jiang2024survey} & Rawte et al. \cite{rawte2023troubling}\\ 
\midrule
Mathematical Formalization  & \textcolor{green}{\ding{51}}& \textcolor{red}{\ding{55}} & \textcolor{red}{\ding{55}} & \textcolor{red}{\ding{55}} & \textcolor{red}{\ding{55}} & \textcolor{red}{\ding{55}} & \textcolor{red}{\ding{55}} & \textcolor{red}{\ding{55}}\\

Mathematical Origins of Hallucinations & \textcolor{green}{\ding{51}} & \textcolor{red}{\ding{55}} & \textcolor{red}{\ding{55}} & \textcolor{red}{\ding{55}} & \textcolor{red}{\ding{55}} & \textcolor{red}{\ding{55}} & \textcolor{red}{\ding{55}} & \textcolor{red}{\ding{55}}\\

Empirical Causes of Hallucinations & \textcolor{green}{\ding{51}} & \textcolor{red}{\ding{55}} & \textcolor{red}{\ding{55}} & \textcolor{green}{\ding{51}} & \textcolor{green}{\ding{51}} & \textcolor{green}{\ding{51}} & \textcolor{red}{\ding{55}} & \textcolor{red}{\ding{55}}\\

Taxonomy of Hallucinations & \textcolor{green}{\ding{51}}  & \textcolor{red}{\ding{55}} & \textcolor{red}{\ding{55}} & \textcolor{green}{\ding{51}} & \textcolor{green}{\ding{51}} & \textcolor{red}{\ding{55}} & \textcolor{red}{\ding{55}} & \textcolor{green}{\ding{51}}\\

Detection of Hallucinations & \textcolor{green}{\ding{51}} & \textcolor{green}{\ding{51}} & \textcolor{green}{\ding{51}} & \textcolor{green}{\ding{51}} & \textcolor{green}{\ding{51}} & \textcolor{red}{\ding{55}} & \textcolor{green}{\ding{51}} & \textcolor{red}{\ding{55}}\\

Evaluation of Hallucinations & \textcolor{green}{\ding{51}} & \textcolor{red}{\ding{55}} & \textcolor{red}{\ding{55}} & \textcolor{green}{\ding{51}} & \textcolor{red}{\ding{55}} & \textcolor{green}{\ding{51}} & \textcolor{green}{\ding{51}} & \textcolor{green}{\ding{51}}\\

Mitigation of Hallucinations & \textcolor{green}{\ding{51}} & \textcolor{green}{\ding{51}} & \textcolor{green}{\ding{51}} & \textcolor{green}{\ding{51}} & \textcolor{green}{\ding{51}} & \textcolor{green}{\ding{51}} & \textcolor{green}{\ding{51}} & \textcolor{green}{\ding{51}}\\

Datasets and Benchmarks & \textcolor{green}{\ding{51}} & \textcolor{red}{\ding{55}} & \textcolor{green}{\ding{51}} & \textcolor{red}{\ding{55}} & \textcolor{green}{\ding{51}} & \textcolor{green}{\ding{51}} & \textcolor{red}{\ding{55}} & \textcolor{red}{\ding{55}}\\

Limitations of Existing Methods & \textcolor{green}{\ding{51}} & \textcolor{green}{\ding{51}} & \textcolor{red}{\ding{55}} & \textcolor{green}{\ding{51}} & \textcolor{red}{\ding{55}} & \textcolor{green}{\ding{51}} & \textcolor{red}{\ding{55}} & \textcolor{red}{\ding{55}}\\

Future Directions & \textcolor{green}{\ding{51}} & \textcolor{green}{\ding{51}} & \textcolor{green}{\ding{51}} & \textcolor{green}{\ding{51}} & \textcolor{green}{\ding{51}} & \textcolor{green}{\ding{51}} & \textcolor{green}{\ding{51}} & \textcolor{red}{\ding{55}}\\


\bottomrule
\end{tabular}
}
\end{table*}

The growing integration of LLMs into societal infrastructures has precipitated a critical research challenge: the emergence of hallucinatory phenomena that exhibit escalating prevalence across both casual applications and high-stakes professional environments \cite{FosterMcBride2024AITRUST}. 
This phenomenon manifests with particular severity in legally sensitive contexts, where LLM-generated outputs may contain substantive factual inaccuracies or systematically misleading interpretations of jurisprudence \cite{buchicchio2024design}. 
Similar challenges emerge across multiple mission-critical domains through distinct mechanistic pathways: 
In healthcare systems, hallucinated diagnostic recommendations may compromise patient safety by deviating from evidence-based clinical guidelines \cite{Yuan2023Large}. 
Financial sector implementations face analogous risks through the generation of non-compliant investment strategies or misrepresentation of market regulations \cite{Chen2024Survey}. 
Educational applications demonstrate particular vulnerability, where hallucinatory content could distort pedagogical frameworks and undermine scholarly credibility \cite{Chen2024AI}. 
Thus, the widespread prevalence of hallucinatory phenomena in LLMs highlights the critical need for a comprehensive survey to thoroughly examine current research advancements and mitigation strategies. 




Existing surveys have extensively documented hallucination phenomena in LLMs as shown in Table~\ref{tab:all_works1}. 
However, three critical limitations persist in existing surveys. 
First, existing surveys generally investigate hallucinations through empirical perspectives, focusing on phenomenological definitions and descriptive analyses of their origins. 
However, an emerging research paradigm — spearheaded by theoretical explorations (through learning-theoretic methodologies \cite{author2024inevitable}\cite{suzuki2025hallucinations}\cite{10.1145/3618260.3649777}) — has initiated rigorous mathematical analysis of hallucination mechanisms.
This analytical shift enables the systematic revelation of fundamental operational principles within hallucinations, which has been largely ignored by existing surveys. 
Second, existing surveys employ broad taxonomies to classify hallucinations, assuming cross-domain uniformity \cite{lin2024towards}. 
However, the manifestations of hallucinations are inherently domain-specific \cite{lavrinovics2025knowledge}. 
For example, in healthcare domains, hallucinations frequently involve inaccurate clinical references or misdiagnostic claims \cite{pal-etal-2023-med}, while in educational contexts, they predominantly stem from distorted representations of conceptual hierarchies or pedagogical frameworks \cite{ahmad2023hallucinations}. 
Such generic taxonomies in existing surveys lead to systematic mischaracterizations of domain-specific hallucinations and impose evaluation criteria that lack contextual granularity, thereby limiting the development of nuanced frameworks for understanding hallucination mechanisms. 
Third, existing surveys primarily focus on summarizing hallucination mitigation techniques while overlooking the constraints or inherent limitations of these solutions \cite{ahmad2023hallucinations}. 
For instance, current approaches fail to address structural weaknesses in mitigation frameworks \cite{zhang2025hallucination}, their heavy reliance on task-specific assumptions \cite{lin2024towards}, and their limited adaptability to dynamic or unpredictable data environments \cite{tonmoy2024comprehensive}. 
The lack of such foundational analysis in survey literature leaves researchers without a comprehensive roadmap for developing next-generation mitigation strategies.

In this survey, 
Our work addresses these gaps through three fundamental contributions. First, we establish the first unified theoretical framework for LLM hallucinations, formally resolves the long-standing research fragmentation in the field.
Second, we reveal why current mitigation approaches are fundamentally constrained by demonstrating the unavoidable nature of hallucinations.
Third, we develop a novel task-aware evaluation taxonomy that systematically links semantic divergence metrics to model architecture properties, enabling precision diagnostics across different NLP tasks. Together, these contributions provide both theoretical foundations and practical tools that advance beyond the current state-of-the-art documented in prior surveys.

This survey aims to conduct an in-depth analysis of hallucinations in large language models (LLMs), focusing on uncovering the underlying mechanisms that drive hallucination generation and examining the latest techniques for their detection, evaluation, and mitigation. The objective is to develop a comprehensive framework for the study of LLM hallucinations, thereby fostering further research in this area. The paper is organized as follows: first, we present the Mathematical Origins, outlining the theoretical foundations; second, we explore the Empirical Causes, identifying the factors contributing to the issue; third, we provide an Evaluation of existing approaches and methodologies; fourth, we discuss Detection techniques for identifying the problem; and finally, we propose Mitigating Methods to address and reduce the impact.

In summary, the primary contributions of this paper to the study of LLM hallucinations are threefold:
\begin{enumerate}
    \item \textbf{Mechanistic Insights into Various Hallucination Types}: This paper offers an in-depth analysis of the mechanisms driving different types of hallucinations, such as factual inaccuracies, unfaithful representations, and logical inconsistencies. By examining the internal architecture and generation processes of LLMs, it identifies the fundamental causes of these hallucinations, thereby laying a theoretical foundation for understanding the phenomenon.

    \item \textbf{Comprehensive Methods for Evaluation, Detection, and Mitigation}: The paper systematically reviews multiple approaches for evaluating and detecting hallucinations, encompassing automatic evaluation metrics, human assessment methods, and specialized detection frameworks. It also proposes mitigation strategies, including Retrieval-Augmented Generation (RAG), the use of knowledge graphs, and prompt engineering, providing concrete solutions to enhance LLM reliability and practical application.

    \item \textbf{Exploration of Future Research Directions}: By analyzing the causes of hallucinations and the constraints of current approaches, this paper identifies gaps in existing research and suggests avenues for future work, such as investigating the relationship between hallucinations and the subspace of true information, as well as the role of confidence calibration in hallucination occurrence. These proposals aim to inspire further academic research in the field.
\end{enumerate}

\newsavebox{\measurebox}
\definecolor{paired-light-blue}{RGB}{198, 219, 239}
\definecolor{paired-dark-blue}{RGB}{49, 130, 188}
\definecolor{paired-light-orange}{RGB}{251, 208, 162}
\definecolor{paired-dark-orange}{RGB}{230, 85, 12}
\definecolor{paired-light-green}{RGB}{199, 233, 193}
\definecolor{paired-dark-green}{RGB}{49, 163, 83}
\definecolor{paired-light-purple}{RGB}{218, 218, 235}
\definecolor{paired-dark-purple}{RGB}{117, 107, 176}
\definecolor{paired-light-gray}{RGB}{217, 217, 217}
\definecolor{paired-dark-gray}{RGB}{99, 99, 99}
\definecolor{paired-light-pink}{RGB}{222, 158, 214}
\definecolor{paired-dark-pink}{RGB}{123, 65, 115}
\definecolor{paired-light-red}{RGB}{231, 150, 156}
\definecolor{paired-dark-red}{RGB}{131, 60, 56}
\definecolor{paired-light-yellow}{RGB}{231, 204, 100}
\definecolor{paired-dark-yellow}{RGB}{141, 209, 49}
\tikzset{%
    parent/.style =          {align=center,text width=1.5cm,rounded corners=3pt, line width=0.3mm, fill=pink!10,draw=pink!80},
    child/.style =           {align=center,text width=2.3cm,rounded corners=3pt, fill=blue!10,draw=blue!80,line width=0.3mm},
    grandchild/.style =      {align=center,text width=2cm,rounded corners=3pt},
    greatgrandchild/.style = {align=center,text width=1.5cm,rounded corners=3pt},
    greatgrandchild2/.style = {align=center,text width=1.5cm,rounded corners=3pt},    
    referenceblock/.style =  {align=center,text width=1.5cm,rounded corners=2pt},
    data/.style =           {align=center,text width=2cm,rounded corners=3pt, fill=paired-light-blue!50,draw=paired-dark-blue!65,line width=0.3mm},
    data_wide/.style =           {align=center,text width=4cm,rounded corners=3pt, fill=paired-light-blue!50,draw=paired-dark-blue!65,line width=0.3mm},   
    data_work/.style =           {align=left, text width=6.5cm,rounded corners=3pt, fill=blue!0,draw=paired-dark-blue!65,line width=0.3mm},  
    model/.style =           {align=center,text width=2cm,rounded corners=3pt, fill=paired-light-orange!50,draw=paired-dark-orange!65,line width=0.3mm},  
    model_wide/.style =           {align=center,text width=4cm,rounded corners=3pt, fill= paired-light-orange!50,draw=paired-dark-orange!65,line width=0.3mm}, 
    model_more/.style =           {align=center,text width=4cm,rounded corners=3pt, fill=paired-light-orange!50,draw=paired-dark-orange!65,line width=0.3mm},
    model_more_left/.style =      {align=left,text width=4cm,rounded corners=3pt, fill=paired-light-orange!50,draw=paired-dark-orange!65,line width=0.3mm},
    model_large_left/.style =      {align=left,text width=6.45cm,rounded corners=3pt, fill=paired-light-orange!50,draw=paired-dark-orange!65,line width=0.3mm},   
    model_work/.style =           {align=left,text width=6.5cm,rounded corners=3pt, fill=red!0,draw=paired-dark-orange!65,line width=0.3mm},
    model_work_left/.style =      {align=left,text width=4cm,rounded corners=3pt, fill=paired-light-orange!50,draw=red!0,line width=0.3mm}, 
    model_work_small/.style =     {align=left,text width=4cm,rounded corners=3pt, fill=paired-light-orange!50,draw=red!0,line width=0.3mm},  
    model_work_small_2/.style =     {align=left,text width=6cm,rounded corners=3pt, fill=paired-light-orange!50,draw=red!0,line width=0.3mm}, 
    pretraining/.style =           {align=center,text width=2cm,rounded corners=3pt, fill= paired-light-green!50,draw=paired-dark-green!75,line width=0.3mm}, 
    pretraining_wide/.style =           {align=center,text width=4cm,rounded corners=3pt, fill= paired-light-green!50,draw=paired-dark-green!75,line width=0.3mm}, 
    pretraining_more/.style =           {align=center,text width=4cm,rounded corners=3pt, fill= paired-light-green!50,draw=paired-dark-green!75,line width=0.3mm},   
    pretraining_work/.style =           {align=left,text width=6.5cm,rounded corners=3pt, fill= cyan!0,draw=paired-dark-green!75,line width=0.3mm},      
    finetuning/.style =           {align=center,text width=2cm,rounded corners=3pt, fill= paired-light-purple!50,draw=paired-dark-purple!75,line width=0.3mm},   
    finetuning_wide/.style =      {align=center,text width=4cm,rounded corners=3pt, fill= paired-light-purple!50,draw=paired-dark-purple!75,line width=0.3mm},   
    finetuning_work/.style =           {align=center,text width=8cm,rounded corners=3pt, fill= paired-light-purple!50,draw= orange!0,line width=0.3mm},        
    inference/.style =           {align=center,text width=2cm,rounded corners=3pt, fill= paired-light-red!35,draw=paired-light-red!90,line width=0.3mm},           
    inference_more/.style =           {align=center,text width=4cm,rounded corners=3pt, fill= paired-light-red!35,draw=paired-light-red!90,line width=0.3mm},
    inference_work/.style =           {align=left,text width=6.5cm,rounded corners=3pt, fill=orange!0,draw=paired-light-red!90 ,line width=0.3mm},     
    application/.style =           {align=center,text width=2cm,rounded corners=3pt, fill= paired-light-yellow!15,draw=paired-light-yellow!90,line width=0.3mm},        
    application_wide/.style =           {align=center,text width=4cm,rounded corners=3pt, fill= paired-light-yellow!15,draw=paired-light-yellow!90,line width=0.3mm},
    application_more/.style =      {align=left,text width=6.5cm,rounded corners=3pt, fill= magenta!0,draw=paired-light-yellow!90,line width=0.3mm},
    application_work/.style =      {align=center,text width=4.5cm,rounded corners=3pt, fill= paired-light-yellow!15,draw= magenta!0,line width=0.3mm},   
}

\begin{figure*}
\centering
\scriptsize
\hspace*{-30pt}
    \begin{forest}
        for tree={
            forked edges,
            grow'=0,
            draw,
            rounded corners,
            node options={align=center,},
            text width=2.7cm,
            s sep=6pt,
            calign=edge midpoint,
        },
        [\textbf{Hallucination} \\ \textbf{Survey} \\ \textbf{Workflow}, fill=gray!45, parent
            [Mathematical Origins, application
                [Undecidability Principles, application_wide
                    [Gödel's Incompleteness Theorems, application_wide
                        [First \& Second Incompleteness Theorem, application_more]
                    ]
                    [Turing's Halting Problem, application_wide
                        [Sequential Processing Without Termination Foresight, application_more]
                    ]
                ]
                [Mathematical Constraints in LLMs, application_wide
                    [Formal System Boundaries, application_wide
                        [Short-Sightedness; Schrödinger's Memory, application_more]
                    ]
                    [Probability Space Constraints, application_wide
                        [Local Optimization with Path Blindness, application_more]
                    ]
                ]
                [Mathematical Inevitabilities, application_wide]
            ]
            [Empirical Causes, pretraining 
                [Data-Related Limitations, pretraining_wide 
                    [Timeliness Problems, pretraining_wide 
                        [(a): Knowledge Lag; (b): Outdated Training Data, pretraining_work
                        ]
                    ] 
                    [Data Quality Issues, pretraining_wide 
                        [(a): Factual Misrepresentations; (b): Online Contamination, pretraining_work
                        ] 
                    ] 
                ] 
                [Model Architecture Constraints, pretraining_wide 
                    [Attention Mechanism Deficiencies, pretraining_wide 
                        [(a): ``Short-Sightedness''; (b): ``Schrödinger's Memory'', pretraining_work
                        ] 
                    ] 
                    [Decoding Strategy Limitations, pretraining_wide 
                        [(a): Local Optimization Problems; (b): Generation Path Blindness, pretraining_work
                        ] 
                    ] 
                ] 
                [Cognitive Processing Barriers, pretraining_wide 
                    [Calibration Challenges, pretraining_wide 
                        [(a): Fine-Tuning Overconfidence; (b): Knowledge Forgetting, pretraining_work
                        ] 
                    ] 
                    [Input Comprehension Difficulties, pretraining_wide 
                        [(a): Cultural Nuances; (b): Multi-Layer Meanings, pretraining_work] 
                    ] 
                ] 
            ]
            [Evaluation, for tree={ data}
                [Metric-Based Evaluation, data_wide
                    [Classification, data_wide
                        [(a): Accuracy; (b): F1 Score, data_work]
                    ]
                    [Generation, data_wide
                        [(a): BLEU; (b): ROUGE; (c): BERTScore, data_work]
                    ]
                    [Confidence, data_wide
                        [(a):ECE; (b): MACROCE; (c): ERCE; (d): IPR and CE, data_work]
                    ]
                ]
                [Benchmark-Based Evaluation, data_wide 
                    [General Domain Benchmarks, data_wide 
                        [(a): HaluEval \cite{li2023halueval}; (b): DEFAN \cite{rahman2024defan}, data_work
                        ] 
                    ] 
                    [Task-Specific Benchmarks, data_wide 
                        [(a): Math \cite{sun2024benchmarking}; (b): Long QA \cite{sachdeva2024fine}, data_work
                        ]
                    ] 
                    [Vertical Domain Benchmarks, data_wide 
                        [(a): Health; (b): Legal; (c): Science \cite{lin2021truthfulqa}, data_work
                        ] 
                    ] 
                    [Method Evaluation Benchmarks, data_wide
                        [(a): Mitigation \cite{chen2024unified}; (b): Internal \cite{su2024unsupervised}, data_work
                        ] 
                    ]
                ]
            ]
            [Detection, for tree={fill=red!45,model}
                [White-Box Model Detection, model_wide 
                    [Embedding-Based Methods, model_wide 
                        [(a): Hidden Layers \cite{zhang2024transferable}; (b): Context MLP \cite{su2024unsupervised}, model_work
                        ] 
                    ] 
                    [Logit-Based Methods, model_wide 
                        [(a): NN Analysis \cite{ji2024llm,liu2024uncertainty}; (b): Token Prob. \cite{kumar2024confidence} (c): Semantic Entropy \cite{kuhn2023semantic,farquhar2024detecting}; (d): Uncertainty Types \cite{xiao2021hallucination}, model_work
                        ]
                    ]
                    [Activation-Based Methods, model_wide 
                        [(a): Penult. Layer \cite{chen2024inside}; (b): Overshadowing \cite{zhang2024knowledgeovershadowingcausesamalgamated} (c): Attention Maps \cite{chuang-etal-2024-lookback}; (d): Graph Segment \cite{nonkes-etal-2024-leveraging}, model_work
                        ]
                    ]
                ]
                [Black-Box Model Detection, model_wide      
                    [Consistency-Based Methods, model_wide
                        [(a): CoT Reasoning \cite{wu2024uncertainty,xue2023rcot}; (b): Response Variation \cite{manakul2023selfcheckgpt,qiu2023detecting,zhang2023sac}; (c): Consistency FT \cite{park2024mitigating}; (d): Layer Analysis \cite{kim2024detectingllmhallucinationlayerwise}, model_work
                        ] 
                    ] 
                    [Confidence-Based Methods, model_wide 
                        [(a): Verbal Uncertainty \cite{lin2022teaching,xiong2023can,yona2024can,zhou2023navigating}; (b): Canary Lookahead \cite{li2024hallucanafixingllmhallucination}; (c): Curiosity Auditing \cite{zheng2025calmcuriositydrivenauditinglarge}, model_work
                        ] 
                    ] 
                    [Auxiliary Models and Tools, model_wide 
                        [(a): LLM Detector \cite{pelrine2023towards,kirstein2024s}; (b): LLM Discussion \cite{chang2024uncovering}; (c): SLM Filtering \cite{hu2024slm}; (d): Knowledge Tools \cite{chen2024unified,deng2024llmgoodpathplanner}, model_work
                        ] 
                    ]
                ]
            ]
            [Mitigating Methods, for tree={inference}
                [Shifting Demand Strategy, inference_more 
                    [Refusal Strategies, inference_more 
                        [(a): Direct Refusal \cite{amayuelas2023knowledge,zhang2024r}; (b): Knowledge Boundary Recognition \cite{cheng2024can}, inference_work
                        ]
                    ] 
                    [Training-time Calibration, inference_more 
                        [(a): Fine-tuning \cite{band2024linguistic}; (b): PPO \cite{xu2024sayself}; (c): Pre-train Feature Recovery \cite{he2023preserving}, inference_work
                        ]
                    ]
                    [Inference-time Calibration, inference_more 
                        [(a): Self-expression \cite{tian2023just,zhou2023navigating}; (b): Few-shot \cite{li2024few}; (c): RAG \cite{ren2023investigating,yin2023large}, inference_work
                        ]
                    ] 
                ]
                [Task Simplification Strategy, inference_more 
                    [Retrieval Augmented Generation, inference_more 
                        [(a): Knowledge Source \cite{anjum2024halo,zhu2024information}; (b): Integration \cite{xu2024unsupervised,qian2024grounding}; (c): Balance \cite{sun2024redeep}, inference_work
                        ]
                    ] 
                    [Knowledge Graph, inference_more 
                        [(a): Structure \cite{guan2024mitigating}; (b): Entity Relations \cite{jiang2022unikgqa,liu2024knowledge}; (c): Multi-hop Reasoning \cite{gao2024two,pusch2024combining}, inference_work
                        ]
                    ] 
                    [Prompt Engineering, inference_more 
                        [(a): CoT \cite{wei2022chain,zhou2022least}; (b): Auto-CoT \cite{zhang2022automatic,press2022measuring}; (c): Self-Critique \cite{mundler2023self,diallo2025iaopromptingmakingknowledge}, inference_work
                        ]
                    ] 
                    [Reflection Methods, inference_more 
                        [(a): Self \cite{zhang2024self,zhang2024selfcontrastbetterreflectioninconsistent,chen2023adaptation}; (b): External \cite{chang2024uncovering,huang2024queryagent,watson2024thingbadquestionh4r}, inference_work
                        ]
                    ] 
                ]
                [Capability Enhancement Strategy, inference_more 
                    [Fine-Tuning Methods, inference_more 
                        [(a): PEFT \cite{hu2022lora,zhao2023knowledgeable,chen2024yes}; (b): Confidence Calibration \cite{wang2024uncertainty}; (c): Knowledge Editing \cite{fang2024alphaedit,orgad2024llms}, inference_work
                        ]
                    ] 
                    [Structural Optimization, inference_more 
                        [(a): Attention \cite{yu2024interpreting,chen2024selfie,verma2024adaptive}; (b): Activation Space \cite{chen2024inside,li2024inference,zhang2024truthx}; (c): Layer Adaptation \cite{liang2024internal,alain2016understanding,wang2022interpretability}, inference_work
                        ]
                    ] 
                    [Decoding Strategies, inference_more 
                        [(a): Context-Aware \cite{shi2023trusting,chuang2024lookback}; (b): Attention Control \cite{zhang2024paying,zhao2024enhancing,gema2024decore}; (c): Contrastive \cite{wang2024mathbb,zhang2023alleviating,chuang2023dola}, inference_work
                        ]
                    ] 
                ]
            ]
        ]
    \end{forest}
    \vspace{-2pt}
    \caption{Overview of the paper workflow.}
    \vspace{-15pt}
    \label{fig:paper_structure}
\end{figure*}
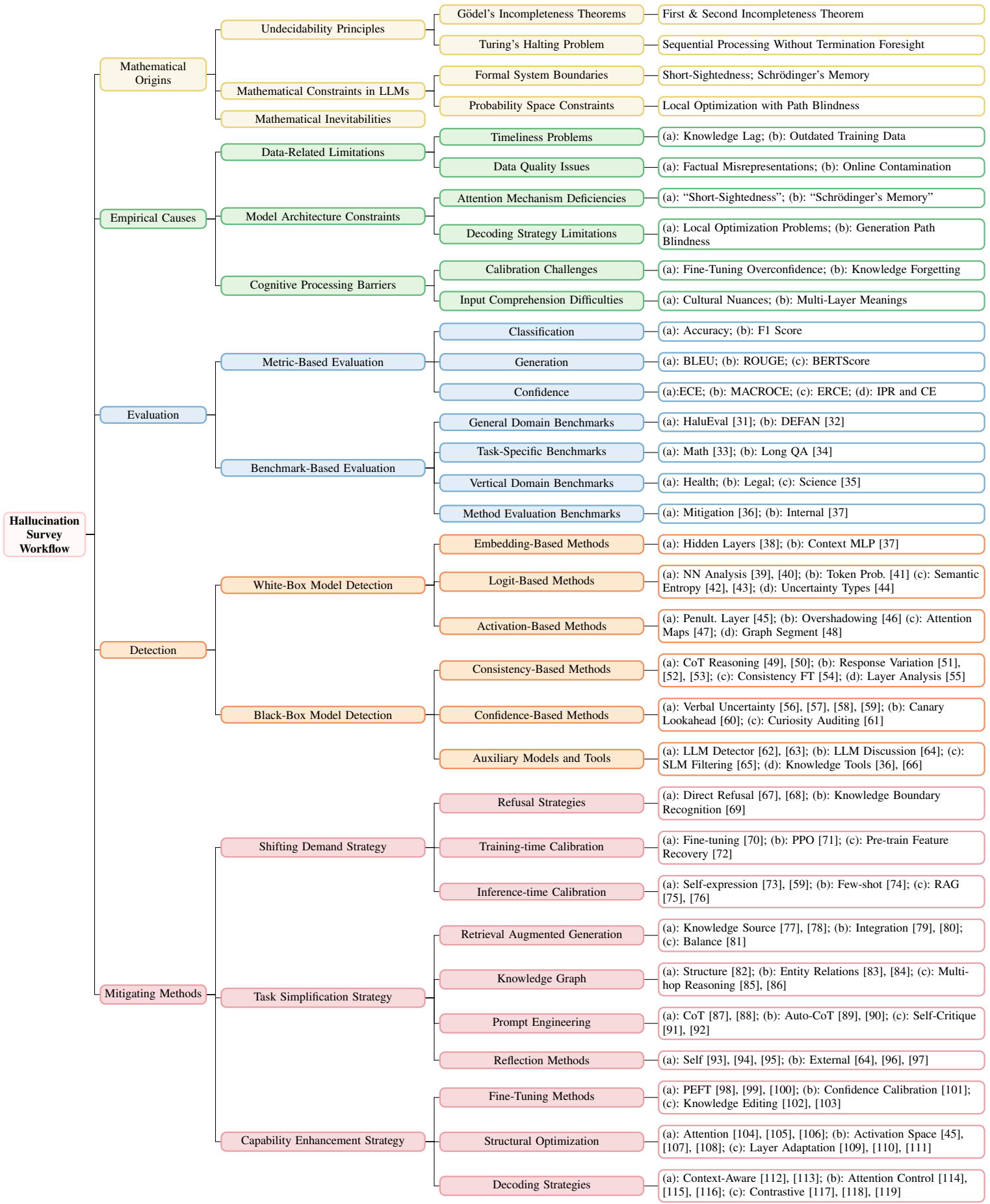

\section{Definitions}

In this section, we have elaborated on the critical concepts of hallucinations in Large Language Models (LLMs), including their formal definitions, primary categorizations—factual, faithfulness, and logical fallacy hallucinations—as well as how these hallucinations arise and evolve through the model's training processes of pre-training and post-training. We also explored the significance of confidence calibration in LLMs and how training phases impact this aspect. This sets a solid groundwork for a deeper understanding and the addressing of hallucination issues. We begin by addressing the issue of hallucinations within large language models.


\subsection{Hallucinations in Large Language Models}

\begin{figure}
\centering
\begin{tikzpicture}[
    scale=0.9,
    query/.style={circle, draw, fill=blue!20, opacity=0.7, minimum size=10pt, inner sep=0pt, font=\footnotesize},
    response/.style={circle, draw, fill=green!70, opacity=0.7, minimum size=5pt, inner sep=0pt},
    hallucination/.style={circle, draw, fill=red!70, opacity=0.7, minimum size=5pt, inner sep=0pt},
    set/.style={rounded corners=3pt, fill=gray!5},
    canonical/.style={dashed, rounded corners=3pt}
]

\draw [set] (-4.5,-2.5) rectangle (4.5,2.5);
\node at (0,2.8) {$\Sigma^*$ (String Space)};

\draw[dotted] (-1.5,-2.5) -- (-1.5,2.5);

\node at (-3,2) {Query Space};
\node at (2,2) {Response Space};

\node[query] (q1) at (-3,1) {$q_1$};
\node[query] (q2) at (-3,0) {$q_2$};
\node[query] (q3) at (-3,-1) {$q_3$};

\draw [canonical, draw=green!50!black] (0,0.7) rectangle (1.5,1.3);
\node[font=\footnotesize] at (0.75,1.5) {$F(q_1)$};

\draw [canonical, draw=green!50!black] (0.5,-0.3) rectangle (2,0.3);
\node[font=\footnotesize] at (1.25,0.5) {$F(q_2)$};

\draw [canonical, draw=green!50!black] (0,-1.3) rectangle (1.5,-0.7);
\node[font=\footnotesize] at (0.75,-0.5) {$F(q_3)$};

\node[response] (r1) at (0.75,1) {};
\node[font=\tiny] at (0.9,1.15) {$r_1$};

\node[response] (r2) at (1.25,0) {};
\node[font=\tiny] at (1.4,0.15) {$r_2$};

\node[response] (r3) at (0.75,-1) {};
\node[font=\tiny] at (0.9,-0.85) {$r_3$};

\node[hallucination] (h1) at (0.05,0.73) {};
\node[font=\tiny] at (0.2,0.88) {$h_1$};

\node[hallucination] (h2) at (3,-0.1) {};
\node[font=\tiny] at (3.15,0.05) {$h_2$};

\node[hallucination] (h3) at (1.4,-1.39) {};
\node[font=\tiny] at (1.55,-1.24) {$h_3$};

\draw[->, blue, thick] (q1) to[bend left=5] (r1);
\draw[->, blue, thick] (q2) to[bend left=5] (r2);
\draw[->, blue, thick] (q3) to[bend left=5] (r3);

\draw[->, red, thick] (q1) to[bend right=5] (h1);
\draw[->, red, thick] (q2) to[bend right=5] (h2);
\draw[->, red, thick] (q3) to[bend right=5] (h3);

\node[font=\footnotesize] at (0,-1.9) {
    \textcolor{blue}{$\rightarrow$} Valid Responses \quad
    \textcolor{red}{$\rightarrow$} Hallucinations
};

\end{tikzpicture}
\caption{Formal Representation of Query-Response Mapping in LLMs}
\label{fig:hallucination-formal}
\end{figure}
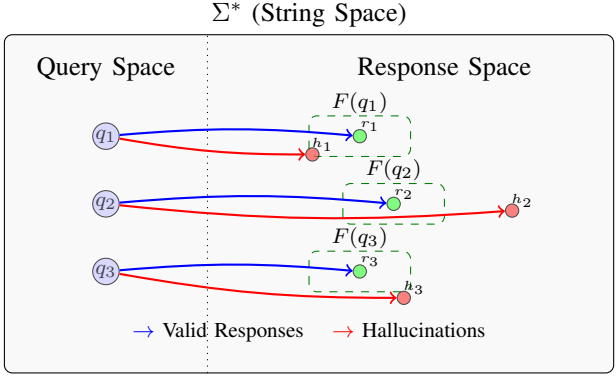

Before formally defining hallucination, it is necessary to establish what constitutes a canonical response in the real world.

Let $\Sigma$ be an alphabet, and denote by $\Sigma^*$ the set of all finite-length strings that can be formed from the characters in $\Sigma$. For a given query $q \in \Sigma^*$, there may exist one or multiple canonical responses that are considered correct or acceptable. We denote by $F(q)$ the set of all canonical responses for query $q$. For some well-defined queries, $F(q)$ may contain exactly one element, while for open-ended or ambiguous queries, $F(q)$ may contain multiple valid responses.
We define the set of all valid query-response pairs as:
\begin{equation}
    \mathscr{S} := \{(q, r): q \in \Sigma^*, r \in F(q)\}.
\end{equation}

Given this framework, we can now define hallucination in the context of large language models (LLMs). When an LLM generates a response $\text{LLM}(q)$ to a query $q$, we say that this response is a hallucination if $(q, \text{LLM}(q)) \notin \mathscr{S}$, or equivalently, if $\text{LLM}(q) \notin F(q)$. In other words, hallucination occurs when the LLM produces a response that is not among the set of canonical responses for the given query.

This definition can be further refined to account for varying degrees of deviation from canonical responses, which we will explore in subsequent sections.

For non–true-false questions, $F(q)$ may contain multiple acceptable answers; that is, a response is considered canonical as long as it meets the criteria of correctness or appropriateness, even if it is not unique. 

Given this framework, we say that an LLM produces a hallucinated response if, for some query \(q \in \Sigma^*\), it generates an output \(\text{LLM}(q)\) such that \(\text{LLM}(q) \notin F(q)\), or equivalently, the pair \((q, \text{LLM}(q)) \notin \mathscr{S}\). This can be formally expressed as:
\begin{equation}
    \exists q \in \Sigma^* \text{ such that } (q, \text{LLM}(q)) \notin \mathscr{S}.
\end{equation}

Alternatively, we can define consistency of an LLM as the property that every generated response belongs to the set of canonical responses for its corresponding query:
\begin{equation}
    \forall q \in \Sigma^*, \; (q, \text{LLM}(q)) \in \mathscr{S},
    \label{eq:consistency}
\end{equation}
and consider any violation of this condition as a hallucination. This dual perspective allows us to analyze hallucination both as an occurrence of incorrect information and as a breach of the consistency requirement.

In practice, hallucinations can manifest in different degrees of deviation from canonical responses. To formalize these variations, we need to consider the structural elements of responses. Let us represent each response as a set of tokens or semantic units, denoted by $\text{tokens}(r)$ for any response $r$.

Given this representation, we can identify two main forms of hallucination:

First, \textbf{partial correctness}: the generated response \(\text{LLM}(q)\) may share some common elements with canonical responses in $F(q)$, i.e.,
\begin{equation}
    \exists r \in F(q) \text{ such that } \text{tokens}(\text{LLM}(q)) \cap \text{tokens}(r) \neq \emptyset.
\end{equation}

Here, the ``overlap'' represents tokens or information units that appear in both the generated response and at least one canonical response. However, despite this partial overlap, if \((q, \text{LLM}(q)) \notin \mathscr{S}\), the response is still considered a hallucination according to our definition.

Second, \textbf{complete divergence}: there is no commonality between the generated response and any canonical response, that is,
\begin{equation}
    \forall r \in F(q), \text{tokens}(\text{LLM}(q)) \cap \text{tokens}(r) = \emptyset.
\end{equation}

It is important to note that the above formalism treats responses in a purely syntactic manner. Hence, if \(\text{LLM}(q)\) differs from all responses in $F(q)$ in form but conveys the same semantic content, our definition would still classify it as a hallucinated response. Determining semantic equivalence, however, typically requires more sophisticated evaluation metrics beyond the scope of this formal definition.

Having established a formal definition of hallucination, we now examine its various manifestations in large language models (LLMs). From a broader perspective, hallucination typically refers to perceptual experiences that arise without external stimuli. Although these perceptions lack any connection to actual external reality, they appear vivid and real to the observer. In the context of LLMs, this concept manifests when models generate content that deviates from canonical responses in $\mathscr{S}$.

Based on the nature of these deviations from $\mathscr{S}$, we categorize hallucinations into three types:
\begin{enumerate}
    \item \textbf{Factual hallucinations}: These occur when an LLM generates content that conflicts with objective facts or known information. Formally, for a factual query $q$ with a well-established set of canonical responses $F(q)$ based on factual knowledge, the model produces $\text{LLM}(q) \notin F(q)$, contradicting established facts.

    \item \textbf{Faithfulness hallucinations}: These arise when the model's output does not accurately reflect or adhere to the user's instructions contained in query $q$. In terms of our framework, even though multiple responses might be factually correct, only those that follow the specific instructions in $q$ belong to $F(q)$. When $\text{LLM}(q) \notin F(q)$ due to not following instructions, we classify this as a faithfulness hallucination.

    \item \textbf{Logical inconsistencies}: These are characterized by internal contradictions within the model's output or errors in logical reasoning. Here, the canonical response set $F(q)$ contains only logically consistent answers, and $\text{LLM}(q) \notin F(q)$ due to logical flaws in the generated content.
\end{enumerate}

A clear understanding of these different types of hallucinations is essential for enhancing the reliability of LLMs. By rigorously defining and categorizing these hallucinations within our formal framework, we can more effectively pinpoint their root causes and develop targeted mitigation strategies. The following sections will provide a detailed examination of each hallucination type, establishing a foundation for addressing these issues.

\subsection{Classification and Definition of Hallucinations in Large Language Models}

\textbf{Factual Hallucinations}

Factual hallucinations occur when an LLM outputs information that is factually incorrect. Such hallucinations can emerge during both pre-training and fine-tuning phases \cite{du2023quantifying}. As will be discussed later, the architecture of Transformer-based models limits their ability to fully internalize and efficiently use factual knowledge \cite{banerjee2024llms, schuurmans2023memory}. Additionally, the quality of the training and fine-tuning datasets plays a critical role in the model's capacity to learn accurate information \cite{edwards2020go}.

Building on our previous formal framework, we now provide a mathematical definition of factual hallucinations:

\begin{itemize}
    \item Let $\Sigma^*$ be the set of all possible queries, as previously defined.
    
    \item For factual queries that have objectively correct answers based on real-world knowledge, we define $F_{fact}(q) \subseteq F(q)$ as the set of responses that are factually correct for query $q$.
    
    \item Let $D$ represent the set of all factual statements about the real world that are objectively true.
    
    \item Let $X \subseteq D$ be the subset of factual knowledge contained in the training dataset of the LLM.
\end{itemize}

Given our established framework, we can now formally define different types of factual hallucinations. For a factual query $q$, we say a factual hallucination occurs when the LLM generates a response $\text{LLM}(q)$ such that $\text{LLM}(q) \notin F_{fact}(q)$ because it contains statements that contradict information in the set of true factual statements $D$. Formally:
\begin{equation}
\begin{split}
    H_F(q, \text{LLM}(q)) \iff \text{LLM}(q) \notin F_{fact}(q), \\
     \exists s \in \text{statements}(\text{LLM}(q)) : s \notin D,
\end{split}
\end{equation}
where $\text{statements}(\text{LLM}(q))$ represents the set of factual assertions contained in the LLM's response, and $H_F$ denotes factual hallucination.

Additionally, we identify a special case called ``procedural hallucination'' ($H_P$), which occurs when all statements in the model's response are factually correct, but the model arrives at these correct statements through flawed reasoning or without proper evidential basis in its training data:
\begin{equation}
\begin{split}
    H_P(q, \text{LLM}(q)) \iff \text{statements}(\text{LLM}(q)) \subseteq D, \\
     \forall s \in \text{statements}(\text{LLM}(q)), \forall x \in X : x \not\models s,
\end{split}
\end{equation}
where $x \not\models s$ indicates that statement $s$ cannot be logically derived from or is not evidentially supported by any training instance $x \in X$. We classify this scenario as a form of hallucination despite the factual correctness of the output because it represents a fundamental failure in the model's reasoning process. When a model generates correct information through invalid inferential pathways or without proper evidential grounding, this ``correctness'' is merely coincidental and not a result of robust knowledge representation or sound reasoning. This phenomenon undermines the trustworthiness of the model, as it suggests that the model's generation process is disconnected from its training data, making its outputs unpredictable and unreliable even when they happen to be factually accurate.

These definitions highlight that hallucinations are influenced by two key factors: (1) the comprehensiveness of the training data $X$ relative to the entire domain of factual knowledge $D$, and (2) the model's ability to accurately learn and retrieve information from $X$ during generation. While regular factual hallucinations involve generating incorrect information, procedural hallucinations represent cases where the model produces factually correct information by chance rather than through proper learning—a phenomenon that remains problematic for model reliability even when outputs appear correct.

In practical terms, factual hallucinations occur when an LLM generates statements that contradict established facts in the set $D$ of true factual knowledge. To illustrate how our formal definition $H_F$ applies in practice, consider the following examples:

Example 1 (Simple factual error): For the query $q$ = ``When did Joe Biden become U.S. President?'', an LLM might generate the response: ``Joseph Robinette Biden Jr. assumed the U.S. presidency in 2020.'' Since the verified information in $D$ indicates that Biden was inaugurated on January 20, 2021, this statement $s \in \text{statements}(\text{LLM}(q))$ satisfies $s \notin D$, making this response a clear instance of factual hallucination.

Example 2 (Fabricated details): For a query about a historical event, an LLM might include specific details that seem plausible but were never documented or verified—such as invented quotes or precise numerical figures for which no records exist. In this case, the generated statements belong neither to $D$ (verified truths) nor to its complement (verified falsehoods), but rather exist in an epistemically uncertain space. Nevertheless, presenting such unverifiable information as fact constitutes a factual hallucination under our definition.

These examples demonstrate how factual hallucinations relate to our previously defined factors: (1) if information about Biden's inauguration date was underrepresented in the training data $X$ compared to the complete factual domain $D$, or (2) if the model failed to accurately retrieve this information despite its presence in $X$, a factual hallucination would occur. This highlights the practical challenges in ensuring factual reliability in LLMs, even for seemingly straightforward factual queries.

\textbf{Faithfulness Hallucinations}
Faithfulness hallucinations occur when an LLM generates responses that do not follow the user's instructions or deviate from the intended constraints in the query. During the pre-training phase, models often have difficulty understanding and accurately responding to user queries. However, techniques like SFT (Supervised Fine-Tuning) and RLHF (Reinforcement Learning from Human Feedback) can help address this issue \cite{lin2024flame, liao2024enhancing}.

Building on our established formal framework, we define faithfulness hallucinations as follows:

\begin{itemize}
    \item For a given query $q \in \Sigma^*$, let $I(q)$ represent the set of instructions or constraints explicitly or implicitly contained in the query.
    
    \item Let $F_{faith}(q) \subseteq F(q)$ be the set of responses that faithfully follow all instructions in $I(q)$.
    
    \item We introduce the concept of atomic propositions, where both queries and responses can be decomposed into fundamental semantic units $\varepsilon_k$. These atomic propositions represent the basic components of meaning that, when combined, constitute the full query or response.
\end{itemize}

Given these definitions, a faithfulness hallucination occurs when, for a query $q$, the LLM generates a response $\text{LLM}(q)$ such that $\text{LLM}(q) \notin F_{faith}(q)$, specifically because it fails to satisfy one or more instructions in $I(q)$. Formally:
\begin{equation}
\begin{split}
    H_{Fh}(q, \text{LLM}(q)) \iff \text{LLM}(q) \notin F_{faith}(q), \\
     \exists i \in I(q) : \text{LLM}(q) \not\models i,
\end{split}
\end{equation}
where $\text{LLM}(q) \not\models i$ indicates that the model's response does not satisfy or follow instruction $i$.

It is important to note that faithfulness hallucinations differ from factual hallucinations in that the generated content may be factually correct ($\text{statements}(\text{LLM}(q)) \subseteq D$) yet still fail to address the query as instructed. This highlights a key distinction in our taxonomy of hallucinations: while factual hallucinations concern the accuracy of information relative to real-world knowledge, faithfulness hallucinations concern the model's ability to properly interpret and follow the user's intentions and constraints.

To formalize faithfulness hallucinations, we can decompose both the query $q$ and the model's response $\text{LLM}(q)$ into sets of atomic propositions. These atomic propositions represent fundamental semantic units or requirements contained in the query and addressed in the response.

Let $\{\varepsilon_k^q\}_{k\in\mathcal I}$ represent the set of atomic propositions derived from query $q$, where each $\varepsilon_k^q$ corresponds to an instruction or requirement in $I(q)$. Similarly, let $\{\varepsilon_k^r\}_{k\in\mathcal R}$ represent the set of atomic propositions contained in the model's response $\text{LLM}(q)$.

A faithfulness hallucination occurs when there exists at least one instruction or requirement from the query that is not addressed or satisfied by any part of the response. Formally:
\begin{equation}
\begin{split}
    H_{Fh}(q, \text{LLM}(q)) \iff \exists \varepsilon^q \in \{\varepsilon_k^q\}_{k\in\mathcal I}, \\\forall \varepsilon^r \in \{\varepsilon_k^r\}_{k\in\mathcal R} : \varepsilon^r \not\models \varepsilon^q,
\end{split}
\end{equation}
where $\varepsilon^r \not\models \varepsilon^q$ indicates that the proposition $\varepsilon^r$ from the response does not satisfy or address the instruction or requirement $\varepsilon^q$ from the query.

In practical terms, a faithfulness hallucination occurs when the LLM's generated response does not meet the requirements of the query or fails to completely reflect the user's intent. For example, if a user asks, ``Please tell me the time Biden assumed office as President and the opponent he defeated.'' consider two responses:

Example 1: ``Biden secured the Democratic presidential nomination in 2019.'' This response is a faithfulness hallucination because it addresses neither of the query's two requirements ($\varepsilon_1^q$: time of assuming office, $\varepsilon_2^q$: opponent defeated). Formally, for all propositions $\varepsilon^r$ in this response, $\varepsilon^r \not\models \varepsilon_1^q$ and $\varepsilon^r \not\models \varepsilon_2^q$.

Example 2: ``Biden assumed the office of U.S. President on January 20, 2021.'' This response partially addresses the query by satisfying $\varepsilon_1^q$ but fails to mention the opponent Biden defeated ($\varepsilon_2^q$). Therefore, it still constitutes a faithfulness hallucination, though to a lesser degree than the first example.

These examples illustrate how faithfulness hallucinations differ from factual hallucinations: the information provided may be factually correct (as in the second example) but still fails to fully address the user's query as instructed.

\textbf{Logical Inconsistencies}

Logical inconsistencies refer to reasoning errors that occur during an LLM's output generation, such as omissions in reasoning steps, topic deviation, or thought drift, which lead to a lack of logical coherence in the generated response. Both the structure and the training process of the model can influence the occurrence of these issues \cite{eisape2023systematic}.

Continuing with our established formal framework, we define logical inconsistencies as follows:
\begin{itemize}
    \item Let $F_{logic}(q) \subseteq F(q)$ be the set of responses that are logically consistent and coherent for query $q$.
    
    \item The response $\text{LLM}(q)$ can be decomposed into a set of atomic propositions $\{\varepsilon_k^r\}_{k\in\mathcal R}$.
    
    \item Additionally, we can extract various text segments $\{s_j\}_{j \in \mathcal{J}}$ from the response, where each $s_j$ is a continuous substring of $\text{LLM}(q)$.
    
    \item We define a contradiction relation $\oplus$ between propositions, where $\varepsilon_i^r \oplus \varepsilon_j^r$ indicates that propositions $\varepsilon_i^r$ and $\varepsilon_j^r$ are logically contradictory or incompatible with each other.
\end{itemize}

Given these definitions, a logical inconsistency hallucination occurs when the model generates a response containing contradictory propositions. Formally:
\begin{equation}
\begin{split}
    H_L(q, \text{LLM}(q)) \iff \text{LLM}(q) \notin F_{logic}(q), \\
     \exists \varepsilon_i^r, \varepsilon_j^r \in \{\varepsilon_k^r\}_{k\in\mathcal R} : \varepsilon_i^r \oplus \varepsilon_j^r,
\end{split}
\end{equation}
where $\varepsilon_i^r \oplus \varepsilon_j^r$ indicates that the propositions $\varepsilon_i^r$ and $\varepsilon_j^r$ derived from the response contradict each other.

Alternatively, logical inconsistencies can also manifest when propositions derived from one part of the response contradict those from another part:
\begin{equation}
\begin{split}
    \exists \varepsilon^r \in \{\varepsilon_k^r\}_{k \in \mathcal{R}},\ \exists s_j \in \{s_j\}_{j \in \mathcal{J}} : \varepsilon^r \oplus \text{prop}(s_j),
\end{split}
\end{equation}
where $\text{prop}(s_j)$ represents the proposition(s) expressed by the text segment $s_j$.

This definition distinguishes logical inconsistency hallucinations from factual and faithfulness hallucinations by focusing specifically on internal contradictions within the response, regardless of its factual accuracy or adherence to the query instructions.

To illustrate logical inconsistency hallucinations, consider the following example. Given a user query $q$ = ``Please describe Biden's achievements as U.S. President.'' an LLM might generate the following response: 

``Biden assumed office as U.S. President in 2020. The following is his record in office. During Trump's term, he implemented a large-scale policy to increase tariffs on China.''

This response contains multiple types of hallucinations that we can analyze using our formal framework:
\begin{enumerate}
    \item \textbf{Factual hallucination} ($H_F$): The statement ``Biden assumed office as U.S. President in 2020'' is factually incorrect, as Biden was inaugurated in January 2021. Thus, there exists a statement $s \in \text{statements}(\text{LLM}(q))$ such that $s \notin D$.

    \item \textbf{Logical inconsistency} ($H_L$): The response contains contradictory propositions: $\varepsilon_1^r$ = ``The following describes Biden's record in office'' and $\varepsilon_2^r$ = ``During Trump's term, he [Biden] implemented a policy to increase tariffs on China.'' These propositions satisfy $\varepsilon_1^r \oplus \varepsilon_2^r$ because they attribute actions during Trump's presidency to Biden, creating a logical contradiction.

    \item \textbf{Faithfulness hallucination} ($H_{Fh}$): The response fails to address the query's instruction to describe Biden's achievements, instead shifting to discussing policies during Trump's term. This means there exists an instruction $i \in I(q)$ such that $\text{LLM}(q) \not\models i$.
\end{enumerate}

This example demonstrates how different types of hallucinations can co-occur within a single response. The boundaries between these categories are not always distinct—factual errors can lead to logical inconsistencies, and both can contribute to unfaithfulness to the query. This interrelationship reflects the complex nature of hallucinations in LLMs, where different types of reasoning failures often manifest simultaneously and reinforce one another, making both detection and mitigation challenging tasks.

\subsection{Training of Large Language Models}

This section delineates the training paradigm of LLMs, encompassing the entire pipeline from pre-training to inference, with particular emphasis on both the pre-training and post-training phases.

\subsubsection{Pre-training}

Pre-training constitutes the foundational phase in the development of LLMs\cite{wang2023gradual}.
During this phase, the model engages in unsupervised learning, processing vast amounts of unlabeled data to autonomously acquire linguistic patterns and features.
This stage is crucial because it allows the model to develop a comprehensive understanding of natural language without needing explicit human-provided labels or task-specific guidelines \cite{sadeq2023unsupervised}. The LLM learns by predicting missing words or sentences in text fragments, a process known as ''masked language modeling.'' By filling in these gaps, the model grasps complex grammatical structures, semantic relationships, and contextual dependencies that underpin human language \cite{wang2023gradual}. The primary objective of pre-training is to establish a broad and robust language representation that serves as a solid foundation for later stages, such as fine-tuning and domain-specific task optimization \cite{lv2020pre}. This extensive pre-training equips the LLM with versatile linguistic skills, improving its performance across various downstream tasks \cite{wang2023gradual}.

\subsubsection{Post-training}

In the post-training phase, the LLM builds upon the general knowledge acquired during pre-training, refining its ability to apply this knowledge through more targeted approaches. This phase involves two key processes: supervised fine-tuning (SFT) and reinforcement learning from human feedback (RLHF).

During SFT, the LLM is trained on a curated dataset with labeled input-output pairs. This process allows the model to adjust its parameters to better align with specific task objectives, improving its performance on structured tasks. By focusing on task-specific examples, the LLM fine-tunes its understanding, effectively applying the linguistic features gained during pre-training \cite{li2024getting}. This stage also helps minimize errors and inconsistencies by ensuring the model's outputs are contextually appropriate and aligned with human expectations.

After SFT, reinforcement learning from human feedback (RLHF) further enhances the LLM’s alignment with human values and preferences \cite{zhang2023wisdom}. In this stage, human reviewers evaluate the model’s outputs based on criteria such as informativeness, coherence, and factual accuracy, providing feedback or scores. This feedback is used to train a reward model, which guides the LLM in optimizing its responses. Proximal Policy Optimization (PPO) is commonly employed to refine the model based on reward signals from human input. RLHF plays a critical role in improving the accuracy, relevance, and user alignment of the LLM’s responses, bridging the gap between its capabilities and human expectations .

Together, SFT and RLHF transform the LLM from a general-purpose language model into a more refined system better equipped for specific applications, ensuring both reliability and practical utility in the generated content\cite{mehta2023sampleefficientreinforcementlearning}.

\textbf{Supervised Fine-tuning}

Supervised Fine-Tuning (SFT) stands as a cornerstone methodology for refining Large Language Models (LLMs), enabling them to harness pre-existing knowledge more adeptly for specialized domains or tasks \cite{teng2024fine}. This enhancement process leverages finely curated instruction-response pairs found within labeled datasets. Initially, the model employs forward propagation to forecast a response, which is subsequently gauged against the target output. The accuracy discrepancy is quantified through loss functions—predominantly cross-entropy loss or the Kullback-Leibler (KL) divergence in cases involving probabilistic distributions. These functions act as evaluative metrics, illustrating the gap between the model's anticipated outcomes and the desired results. Following this assessment, an iterative adjustment cycle commences, facilitated by backpropagation; the model's weights undergo meticulous recalibration to minimize identified discrepancies and fortify alignment with specified instructions \cite{teng2024fine}.

However, introducing novel knowledge during SFT, distinct from the initial pre-training phase, can instigate a series of complexities. Chief amongst these is the phenomenon known as catastrophic forgetting—the tendency of neural networks to swiftly assimilate new information at the expense of previously acquired skills \cite{luo2023empirical}. This is compounded by confidence calibration issues, manifesting as erratic certainty estimations in the model's outputs(discussed in detail in Section 3). Such anomalies underscore the need for advanced mitigation strategies.  

\textbf{Reinforcement Learning from Human Feedback}

Since the introduction of the classic three-stage training model—Supervised Fine-Tuning (SFT) + Reward Model (RM) + Proximal Policy Optimization (PPO)—the PPO + RLHF (Reinforcement Learning from Human Feedback) approach has become the standard in training\cite{ouyang2022training, liu2020learning}.

Reinforcement Learning with Human Feedback (RLHF) initiates by conditioning a reward model on a substantial human-labeled dataset, equipping it to evaluate Large Language Model (LLM) outputs against human preferences and societal values. Building on this foundation, Proximal Policy Optimization (PPO) assumes a pivotal role. This advanced reinforcement learning algorithm excels in optimizing complex decision-making processes by maintaining a strategic equilibrium between exploration and exploitation \cite{mukobi2023superhf}. PPO ensures that the policy function is updated in a controlled manner, avoiding drastic alterations that could jeopardize learned behaviors . Its clipped objective functions safeguard against performance degradation, promoting steady improvement. Moreover, PPO's use of importance sampling optimizes learning efficiency from multiple episodes without compromising stability. The integration of PPO empowers the reward model to shift from passive observation to active participation in the training process, replacing direct human oversight \cite{zheng2023secrets}. Post pre-training and supervised fine-tuning, this enhanced training protocol enables the LLM to absorb human preferences and values through positive reinforcement \cite{gorbatovski2024reinforcement}. Consequently, the LLM's response quality is significantly enhanced, yielding detailed and high-quality content while markedly reducing the likelihood of harmful, misleading, or irrelevant outputs. The strategic application of PPO within RLHF guarantees that the LLM's outputs evolve continuously towards not just matching but surpassing human expectations across various dimensions, including relevance, accuracy, and adherence to ethical norms. Through judiciously balanced exploration and exploitation, along with the incorporation of human feedback through the reward model, RLHF-equipped LLMs emerge as powerful tools capable of engaging meaningfully with users, reflecting nuanced human values and preferences in every interaction \cite{havrilla2024teaching}. 

Although RLHF aims to guide the LLM to generate comprehensive responses, similar to supervised fine-tuning, the LLM often attempts to provide an answer even when faced with questions beyond its knowledge. This can result in overconfidence in its outputs and the generation of hallucinations. A more detailed exploration of this issue will be discussed in Section 3.

\subsection{Confidence Calibration}

The confidence calibration of a Large Language Model refers to how well its predicted probabilities align with actual outcomes, reflecting the model’s self-awareness and understanding of its knowledge boundaries \cite{xiong2023can}. A well-calibrated model produces output probabilities that closely correspond to the real-world likelihood of events \cite{he2023preserving}. In practical terms, if the model predicts a certain event with a probability of $X\%$, the actual occurrence of that event should approximately match this confidence level. This ensures the model’s predictions are not only accurate but appropriately tempered with uncertainty.

However, while LLMs often maintain good calibration during initial training, this balance frequently becomes compromised in the post-training stage. During post-training fine-tuning, particularly when using reinforcement learning from human feedback (RLHF) or supervised fine-tuning (SFT), the model is encouraged to provide detailed and definitive answers, which can unintentionally affect its ability to gauge certainty accurately \cite{pikus2023baseline}. As a result, when an LLM expresses high confidence during inference, its actual accuracy may fall significantly below its stated confidence level. This misalignment leads to overconfidence, where the model's self-reported certainty does not reflect its true likelihood of being correct \cite{stengel2024lacie}.

Overconfidence can have significant consequences. Although post-training processes are intended to reduce faithfulness hallucinations by aligning responses more closely with desired behavior, this often comes at the cost of proper confidence calibration. Consequently, this imbalance may increase the likelihood of factual hallucinations—instances where the model confidently generates incorrect information \cite{xu2024sayself,bouyamourn-2023-llms}.

The following sections will explore various methods for assessing confidence calibration in LLMs, examining both theoretical foundations and practical applications. By understanding these techniques, we aim to highlight approaches that can improve calibration, reduce overconfidence, and enhance the model’s reliability and overall performance.

\section{Causes of Hallucinations}

Based on the above definitions, this section aims to explore the theoretical causes of hallucinations in LLMs.

Xu \textit{et al.} \cite{xu2024hallucination} leveraged learning theory to demonstrate that LLMs are inherently incapable of learning all computable functions, leading to the inevitable presence of hallucinations. Banerjee \textit{et al.} \cite{banerjee2024llms} further investigated the issue by constructing a series of self-referential paradoxes based on the Turing machine's halting problem, revealing the fundamental reasons why hallucinations arise during both the training and inference stages of LLMs.

This section will provide a detailed mathematical analysis of the principles underlying factual hallucinations, faithfulness hallucinations, and logical inconsistencies. This theoretical exploration will lay the groundwork for assessing the effectiveness of mitigation strategies for hallucinations and will outline the macro-level manifestations of hallucinations across different stages of LLM development and use.

\subsection{Factual Hallucinations}

Factual hallucinations are relatively straightforward to understand, as they stem from two main sources: errors or gaps in the training data and the inherent limitations of large language models. Below, we analyze how incorrect information in the training data leads to factual hallucinations.

Let \( U \) denote the dataset that contains all knowledge in the real world, and let \( X \) be the dataset used to train the LLM. Since no training dataset can fully capture all knowledge, we have $X \subset U $. We further divide \( U \) into \( U_t \), the correct subset of real-world knowledge, and \( U_f \), the incorrect subset, such that \( U = U_t \sqcup U_f \). Similarly, the training data \( X \) consists of a correct portion \( X_t \) and an incorrect portion \( X_f \), giving \( X = X_t \sqcup X_f \). Now, consider a knowledge item \( n' \) that appears only in the erroneous subset of the training data, i.e., \( n' \in X_f \) and \( n' \notin X_t \). When querying about \( n' \) with a query \( q' \), the LLM, having been trained on incorrect information, will be unable to provide an accurate response:

$$
\forall v \in X_f, v \notin X_t: v \not \Rightarrow n \in U_t
$$

In other words, a model trained on false information will be unable to generate correct responses, directly leading to factual hallucinations.

Next, we examine the inherent limitations of LLMs, which imply that even with a complete training dataset, the LLM may still produce factual hallucinations. This limitation is grounded in Gödel's incompleteness theorem, which asserts that any sufficiently powerful formal system cannot be both complete and consistent.

LLMs, particularly those based on Transformer architectures, possess significant expressive power. According to Gödel’s theorem \cite{raatikainen2013godel}, this means that there will always be statements generated by the LLM that cannot be verified. Even if an LLM produces an answer that appears correct, the underlying process of fabrication may itself be a factual hallucination. This can be represented as follows (using the notation from Section 2.2):

$$
\exists \varepsilon^w \in \{\varepsilon_k^w\}_{k \in \mathcal{W}}, \forall v \in X \ \text{s.t.} \ v \not\Rightarrow \varepsilon^w
$$

This expression indicates that due to the model’s inherent incompleteness, its generated outputs will inevitably include unverifiable knowledge, rendering factual hallucinations unavoidable.

\subsection{Faithfulness Hallucinations}

Imagine constructing an ideal large language model that could provide accurate and coherent answers to any human query \( q \). Achieving this would require meeting three key prerequisites: first, the model must accurately understand human queries; second, it must be able to clearly delineate the boundaries of knowledge, distinguishing correct, incorrect, and uncertain information; and third, it must be capable of predicting and selecting the most appropriate response. While these challenges can be examined from a computational theory standpoint, the essence of faithfulness hallucinations is not merely due to technical limitations but is rooted in deeper issues of natural logic.

According to principles like Gödel’s incompleteness theorem \cite{raatikainen2013godel} and the Turing halting problem \cite{rogers1987theory}, any sufficiently complex system will inherently contain statements that are undecidable—neither provable nor disprovable. For such statements, no computational model can provide an absolutely correct answer based on predefined rules. This mathematical limitation implies that certain propositions are fundamentally unresolvable. Therefore, regardless of any improvements made to the architecture or datasets of LLMs, there will always be propositions whose truth or falsehood remains indeterminable, inevitably leading to hallucinations.

We can mathematically demonstrate that the three prerequisites above are inherently unsolvable, making hallucinations unavoidable.

To begin, we can unify the first two prerequisites into a single problem, as both require a critical step—accurate ``intent classification.'' For the first prerequisite, intent classification allows the model to grasp the true meaning of a user query. For the second prerequisite, successful intent classification is necessary for retrieving the most relevant information from the knowledge base. Hence, if we show that perfect intent classification is unattainable, we demonstrate inherent flaws in both prerequisites.

Assume we have an ideal LLM Turing machine \( T \) that can flawlessly discern intent and retrieve the most relevant information. Consequently, there must be a decider \( D_T \) capable of determining, within finite time, whether any input pair \( \langle T, w \rangle \) corresponds to the correct intent.

Now, consider the ``acceptance problem,'' which involves determining whether a Turing machine \( M \), given an input string \( w \), accepts \( w \):
$$
L_a = \{ \langle M, w \rangle : M \text{ accepts } w \}
$$

We construct a decider \( D_M \) that operates as follows:
\begin{itemize}
    \item If \( D_T \) accepts, indicating that the LLM Turing machine \( T \) has correctly classified the intent for \( w \), then \( D_M \) accepts.

    \item If \( D_T \) rejects, indicating that \( T \)'s intent classification is irrelevant, then \( D_M \) rejects.
\end{itemize}

However, the ``acceptance problem'' is known to be undecidable, creating a contradiction in our construction. This demonstrates that perfect intent classification is, in principle, impossible, implying that the first two prerequisites cannot be fully met, leading to hallucinations.

Moreover, it has been shown that the halting problem for LLMs is undecidable, meaning that an LLM cannot predict in advance how many tokens it will generate. This implies that the LLM cannot preemptively decide when to stop generating, theoretically allowing it to produce an infinite sequence of tokens.

For a given query \( q \), generating a correct and coherent response \( w \) requires the model to consider all possible outcomes \( w_k \in \Sigma^* \), where \( \Sigma^* \) is the set of all potential responses, and select the optimal response:
$$
w_m = \max_{k \in \mathcal{A}} P(w_k)
$$

However, since 
$$
|\Sigma^*| = \infty
$$
the possible sequences are infinite, making it impractical for the model to explore all potential outputs. Therefore, the model cannot compute the joint probability of all possible token sequences, rendering the third prerequisite unattainable.

In conclusion, each of the three prerequisites is constrained by problems of undecidability, making faithfulness hallucinations an inevitable outcome. The underlying cause of these hallucinations is not tied to specific design features of LLMs or the limitations of their datasets but rather stems from deeper mathematical and logical issues inherent to undecidability.

\subsection{Logical Inconsistencies}

Gödel’s second incompleteness theorem states that if a formal system \( S \) contains a sufficiently complex arithmetic system and is consistent, it cannot prove its own consistency within its own framework:
$$
S \not\Rightarrow \text{Con}(S)
$$

This theorem reveals the inherent limitations of formal systems: even if a system is assumed to be free from contradictions, it cannot internally verify its own consistency. LLMs, as complex formal systems, inherit this limitation—they are unable to verify the consistency or logical correctness of their own generation processes.

The text generation process of an LLM unfolds step by step, with each token generated conditionally based on the previous context using conditional probabilities:
$$
P(w_i | w_1, w_2, \dots, w_{i-1})
$$

Since an LLM does not have foresight of the complete logical structure of the entire sequence until generation is complete, it is unable to check for logical consistency in real-time. While each step may be a reasonable inference based on local context, the cumulative result of the generation can exhibit logical inconsistencies and biases. This limitation is inherent to the generation process of LLMs, which lack the capacity to maintain global logical coherence throughout the sequence.

Furthermore, due to the undecidability of the halting problem, LLMs cannot determine in advance when they should stop generating tokens. For any given input, the generation sequence could theoretically be infinite:
$$
|\Sigma^*| = \infty
$$

This means that when handling complex queries, an LLM is unable to explore all possible generation paths exhaustively or determine an appropriate stopping point, increasing the risk of logical errors in the output.

In summary, the underlying mechanisms of logical inconsistencies in LLMs stem from three main factors:
\begin{enumerate}
    \item \textbf{Inability to Self-Validate Consistency}: According to Gödel’s second incompleteness theorem, LLMs are incapable of proving the logical consistency of their own generation process. Even though each token generation step may appear coherent, the LLM cannot guarantee that the overall sequence is contradiction-free.
    $$
    LLM \not\Rightarrow \text{Con}(LLM)
    $$

    \item \textbf{Limitations of Sequential Generation}: LLMs generate content sequentially based on conditional probabilities, leveraging context at each step without a mechanism for ensuring global logical coherence. Moreover, the use of locally optimal strategies (such as greedy algorithms) may not lead to a globally coherent result, causing potential logical inconsistencies in the overall output.
    $$
    P(w) = P(w_1) P(w_2 | w_1) \cdots P(w_n | w_1, w_2, \cdots, w_{n-1})
    $$

    \item \textbf{Undecidability of Halting}: Because of the halting problem’s undecidability, LLMs are unable to determine in advance when the generation should end, which can lead to the creation of infinite token sequences or uncontrolled logical drift, thereby raising the likelihood of logical errors.
    $$
    |\Sigma^*| = \infty
    $$
\end{enumerate}
    
Therefore, logical inconsistencies in LLMs are fundamentally unavoidable, stemming from both the intrinsic limitations of formal systems and the challenges posed by sequential generation and undecidability issues.

\subsection{Manifestations of Hallucinations}

This section provides an overview of the three principal categories of hallucinations—Factual Hallucinations, Faithfulness Hallucinations, and Logical Inconsistencies.
It highlights the stages at which each type of hallucination is likely to occur and their macro-level manifestations. 
By scrutinizing the root causes of these hallucinations, this analysis serves as a foundation for proposing targeted mitigation strategies.

\subsubsection{Timeliness and Data Contamination}

Training large language models (LLMs) necessitates the acquisition of extensive datasets. As elucidated by Kaplan \textit{et al.} \cite{kaplan2020scaling}, model performance is profoundly contingent on both the model’s parameterization and the scale of its training corpus—with contemporary LLMs being trained on datasets comprising in excess of 10 trillion tokens. Nevertheless, even with access to such a vast informational reservoir, LLMs confront two formidable challenges: the inherent latency in acquiring up-to-date knowledge and the pervasive issue of data contamination.

In rapidly evolving domains—such as emerging technologies, regulatory frameworks, and real-time news—developments occurring subsequent to the model’s last training update remain unincorporated, resulting in an inevitable lag in current knowledge. Concurrently, the integrity of training data poses a significant challenge; misrepresentations of historical facts, annotation inaccuracies, and extraneous noise from online sources may be inadvertently assimilated during training. This contamination can embed misinformation within the model, thereby precipitating factual hallucinations.

Furthermore, when addressing open-ended tasks, LLMs tend to generate responses characterized by heightened uncertainty and creative divergence from standardized outputs, as noted by Ji \textit{et al.} \cite{ji2023survey}. Although this propensity for originality aligns with certain application directives, it inadvertently fosters conditions conducive to deviations in fidelity. The resultant leapfrog thinking may prompt the model to bypass critical reasoning stages, thus engendering logical inconsistencies or fallacies.

While the incorporation of new information to address timeliness has yielded promising preliminary results, the enhancement of training data quality remains a complex and critical research challenge that warrants further innovative investigation.

\subsubsection{Calibration Challenges}

Section 2.4 delved into the critical issue of calibration in LLMs, particularly overconfidence exacerbated by supervised fine-tuning and reinforcement learning from human feedback post-training. These methodologies, while enhancing an LLM's ability to provide detailed responses to unfamiliar queries, inadvertently stoke the fires of factual hallucinations, a perilous byproduct of enhanced confidence without corresponding accuracy.

gekhman
\textit{et al.}
\cite{gekhman2024does} indicates that  fine-tuning (FT) predominantly hones an LLM's skill in leveraging existing knowledge more effectively, rather than acquiring new knowledge. Intriguingly, LLMs accrue new knowledge at a slower rate than they forget established facts—a dynamic that linearly amplifies the risk of generating hallucinations. Yet, post-pretraining models often fall short of domain-specific requirements or the ability to consistently produce safe, human-beneficial responses, necessitating the incorporation of new knowledge—a compromise between performance and knowledge stability.

Though the calibration quandary post-training remains a challenging hurdle, a promising avenue lies in embedding calibration objectives during the pretraining phase, preemptively averting the need for corrective measures post-training. In the near term, retraining LLMs to recapitulate pretraining features could partially alleviate the overconfidence issue, a strategy that aims to recalibrate models by revisiting foundational knowledge\cite{he2023preserving}. This approach, among other calibration techniques, will be elaborated upon in \textbf{Section 4}, where we explore the latest advancements and research initiatives dedicated to enhancing LLM calibration and mitigating the risks associated with overconfidence and factual hallucinations.

\subsubsection{The “Short-Sightedness” Effect of Attention}

Within the domain of comprehensive context assimilation, Large Language Models (LLMs) grapple with attention deficits,evidenced by the neglect or erosion of initial data points in protracted textual inputs or successive conversational rounds. This occurrence coincides with the genesis of factual discrepancies, faithfulness illusions, and logical aberrations. Specifically, amidst prolonged interactive exchanges, LLMs exhibit a pronounced bias toward immediate exchanges, relegating formative guidelines and critical nuances to the fringes of recall. It should be emphasized that this behavior is not due to deliberate oversight, but rather to intrinsic constraints within their self-attention architectures\cite{wang2024schrodinger}.This cognitive narrowness precipitates generative outputs marred by misinformation, deviation from intended objectives, and internal contradictions.

Superficially, the operational dynamics of LLM memory may mirror those of human short-term memory; however, the true essence reveals a distinct paradigm, aptly characterized by the concept of 'Schrödinger’s Memory',a descriptor coined to encapsulate this elusive trait. Essentially, the model's mnemonic capability operates on a conditional basis, coming alive specifically upon invocation. Thus, the evaluation of memory retention hinges exclusively on the model's responsive output post-query, rendering its status outside of this interaction a quantum-like superposition of presence and absence\cite{wang2024schrodinger}. This unorthodox functionality unveils the frailties of extant attention paradigms, notably in navigating dialogues that traverse multiple themes and extended temporal sequences.

Faced with these limitations, academic pioneers are ardently advancing innovations centered on attention mechanism optimization, geared towards fortifying contextual apprehension. These strides aspire to guarantee the delivery of responses that are meticulously accurate and seamlessly integrated across multi-stage dialogical encounters. In the forthcoming section dedicated to Research Developments (Section Four), we shall immerse ourselves in the vanguard of contemporary scholarly endeavors, tackling these impediments directly. Expect an insightful exploration into novel approaches poised to elevate LLM proficiency in handling intricate and multifaceted dialogue contexts.

\subsubsection{Deeper Challenges in Input Understanding}

LLMs perpetually wrestle with intricate linguistic finesse and cultural subtleties, inevitably encountering interpretive hurdles \cite{li2024llms}. When faced with text rich in polysemous terms, metaphors, or culturally embedded references, the LLM's comprehension often reaches its limit, leading to misinterpretations or ``fidelity hallucinations.''
Consider Deng Ai, a stammering military strategist from the Wei kingdom during the Three Kingdoms period, as a case in point. Under pressure or when introducing himself, Deng Ai stuttered over his name, ``Ai... Ai...''. The emperor jestingly inquired about the number of ``Ai's,'' to which Deng Ai ingeniously retorted ``Phoenix, Oh Phoenix, and thus only one phoenix.'' While an LLM can grasp the symbolism and nobility associated with the phoenix, Deng Ai's response did more than address the literal query about his name; it also paralleled himself to the mythical bird, highlighting his exceptional abilities and uniqueness. Yet, there is an additional layer of meaning. 

The LLM successfully deciphers both the surface meaning and the implied layers of this statement. The phoenix, unique and exalted, allows Deng Ai's reply to address the monarch's overt inquiry about his name while simultaneously drawing a parallel between himself and the noble bird, underscoring his own remarkable talent and unparalleled nature. However, embedded within this exchange lies a tertiary depth of meaning. The phoenix, selective and regal, rests only on the phoenix tree, a metaphor for how the wise choose their companions or leaders. This subtle appreciation of the emperor's wisdom is a complexity that an LLM struggles to decode.

Deng Ai's retort, succinct yet rich in meaning, demands that the model connect historical lore with deeper connotations, a formidable challenge for even the most sophisticated LLM. However, this presents fertile ground for enhancing LLM's interpretative prowess. By augmenting training datasets with high-entropy content, the model's response to user inquiries is fortified \cite{bai2024coig}. This not only bolsters the model's versatility in understanding user needs but also paves the way for a broader spectrum of applications at the convergence of literature and technology. This approach widens the LLM's horizon, enabling it to navigate the rich tapestry of human language and cultural intricacies with greater dexterity.

\subsubsection{Blindness in the Generation Path}

Present decoding strategies including greedy algorithms, Beam Search, and Top-k, Top-p sampling techniques, ensure solely the local quality of generation without guaranteeing a globally optimal path. Considering the greedy algorithm as an exemplar: under the assumption that reduced total output uncertainty signifies superior output quality, we designate the output sequence as \(\mathcal{S}=(x_1,x_2,...,x_T)\), where \(x_t\) stands for the t-th token. In regard to sequence \(\mathbf{x}\), \(P(x_t|\mathbf{x}_{<t})\) symbolizes the probability distribution for producing subsequent token \(x_t\), conditioned upon the prefix \(\mathbf{x}_{<t}=(x_1,x_2,...,x_{t-1})\). At every instant, the greedy algorithm chooses the most probable token as output, expressed as: \(x_t^*=\underset{x}{\mathrm{argmax}} P(x|\mathbf{x}^*_{<t})\); consequently, the sequence \(\mathcal{S}^*=(x_1^*,x_2^*,...,x_T^*)\) embodies the outcome of the greedy strategy.

Shannon's Entropy quantifies randomness in outcomes. Given a random variable \(X\), its entropy is delineated as:

\[ H(X)=-\sum_x P(x)\log P(x) \]

The entropies pertaining to sequences \(\mathcal{S}\) and \(\mathcal{S}^*\) are then computed as follows:

\[ H(\mathcal{S}) =-\sum_{t=1}^{T} P(x_t|\mathbf{x}_{<t})\log P(x_t|\mathbf{x}_{<t}) \]

\[H(\mathcal{S}^*)= -\sum_{t=1}^{T^*} P(x_t^*|\mathbf{x}_{<t})\log P(x_t^*|\mathbf{x}_{<t}) \]

Should an output sequence \(\mathcal{S}\) manifest such that \(H(\mathcal{S}) < H(\mathcal{S}^*)\), akin to an edge scenario wherein \(\mathcal{S}=(0.4,1,1,1)\) juxtaposed against \(\mathcal{S}^*=(0.6,0.7,1)\), notwithstanding the superior quality of output in \(\mathcal{S}\), Language Model (LM) outputs invariably favor \(\mathcal{S}^*\). Methods like Beam Search and Top-k, Top-p sampling, while exploring a wider spectrum of possibilities during response formulation, skirt around the central challenge.
Before finalization, LM models remain oblivious to output quality, termination points, and logical consistency. The prompt engineering approach highlighted in Section 4.3 such as the Chain of Thought (CoT), significantly augments LM responses due to their facilitation of rudimentary comprehension regarding these pivotal aspects, thus mitigating the inherent randomness in the decoding phase.

Beyond prompting techniques, forthcoming studies may delve into novel decoding methodologies, endeavoring to surmount the innate constraints of the prevailing decoding patterns. Nonstandard approaches, for instance, back-to-front decoding, might offer fresh perspectives on untapped potentials for optimization within the context of decoding procedures.

\section{Methods for Detecting and Mitigating Hallucinations}

The section delves into methodologies for addressing hallucinations in LLMs, encompassing detection and mitigation strategies. Detection techniques are split into automatic and manual categories, each serving distinct purposes and offering unique insights. Automatic detection, notably, relies on metric-based and benchmark-based assessments. Metric-based evaluation employs accuracy, the F1 score, Precision, Recall, and metrics like BLEU, ROUGE, and BERTScore for quantifying output fidelity and capturing nuances beyond mere lexical overlap. Benchmark-based evaluation targets domain-specific content, aiming to gauge truthfulness and combat unverifiable information. Manual evaluation complements these efforts by incorporating expert judgment and nuanced understanding, although it faces limitations related to cost, duration, and potential bias. Together, these methods form a comprehensive toolkit for identifying and correcting hallucinatory issues in generative AI systems.

\subsection{Evaluation of Hallucinations}

Hallucination evaluation methods can be broadly categorized into two types: automatic and manual evaluations. Automatic evaluation is further divided into metric-based and benchmark-based approaches.

\subsubsection{Automatic Evaluation}

Automatic evaluation uses predefined quantitative metrics or benchmarks to assess how closely generated content matches actual or expected results. The primary advantages are efficiency and ease of use. Automatic evaluations are typically divided into two types: metric-based and benchmark-based.

\textbf{Metric-Based Evaluation}

Metric-based evaluations are suitable for large-scale datasets, providing rapid feedback and strong generalizability, as they usually do not depend on any specific dataset. Below are commonly used evaluation metrics and their principles.

Classification Tasks: Metrics for classification tasks are especially suitable for datasets with true/false or multiple-choice questions, and they are primarily applied to detect factual hallucinations. Common metrics include accuracy and the F1 score.

Accuracy is a fundamental metric for evaluating classification models, measuring the overall proportion of correctly classified samples:
$$
Accuracy = \frac{TP + TN}{TP + TN + FP + FN}
$$
where:
\begin{itemize}
    \item \textbf{True Positive (TP)}: The number of samples correctly predicted as positive that are indeed positive.

    \item \textbf{True Negative (TN)}: The number of samples correctly predicted as negative that are indeed negative.

    \item \textbf{False Positive (FP)}: The number of samples incorrectly predicted as positive, which are actually negative.

    \item \textbf{False Negative (FN)}: The number of samples incorrectly predicted as negative, which are actually positive.
\end{itemize} 

Although accuracy is straightforward to compute, its effectiveness can be limited when dealing with imbalanced class distributions. For instance, in a dataset with 100 samples where 90 are positive and 10 are negative, a model that predicts all samples as positive would achieve 90\% accuracy. However, this high accuracy fails to capture the model's true classification capabilities. Therefore, the F1 score is often used as a complementary metric to evaluate performance more accurately.

The F1 score is the harmonic mean of precision and recall, making it particularly useful for imbalanced datasets. It aims to balance precision and recall, ensuring that performance is not skewed by the disproportionate size of any one class.
$$
F1 = \frac{2}{\frac{1}{\text{Precision}} + \frac{1}{\text{Recall}}}
$$

where:

\textbf{Precision} measures the proportion of correctly predicted positive samples out of all samples predicted as positive:
$$
Precision = \frac{TP}{TP + FP}
$$

\textbf{Recall} measures the proportion of actual positive samples that are correctly predicted as positive:
$$
Recall = \frac{TP}{TP + FN}
$$

The F1 score, by harmonizing precision and recall, offers a balanced and accurate evaluation of model performance, particularly when dealing with imbalanced datasets, making it effective for identifying factual hallucinations.

Generation Tasks: Unlike the clear-cut evaluation of factual hallucinations, assessing faithfulness hallucinations and logical inconsistencies is more complex. Faithfulness hallucinations are often evaluated in generation tasks by measuring the divergence between the actual output and the expected output. Common evaluation metrics for this purpose include BLEU, ROUGE, and BERTScore.

BLEU originated in machine translation and assesses translation quality by comparing the n-gram overlap between the system-generated translation and reference translations. In hallucination evaluation, BLEU is adapted to measure the similarity between the LLM’s output and the user-prompted expected output.
$$
BLEU = BP \cdot \exp \left( \sum_{n=1}^{N} w_n \log p_n \right)
$$
where \(BP\) is the brevity penalty, \(w_n\) are weights, and \(p_n\) is the n-gram precision. 

ROUGE evaluates the quality of text summarization by comparing the overlap between the generated summary and the reference summary. It includes several variants like ROUGE\_N, based on n-grams, and ROUGE-L, based on the longest common subsequence.

For example, ROUGE-L calculates the overlap by measuring the longest common subsequence (LCS) between the generated text and the reference:
$$
ROUGE_L = \frac{(1 + \beta^2) R_{lcs} P_{lcs}}{R_{lcs} + \beta^2 P_{lcs}}
$$
where \(R_{lcs}\) is recall, defined as \(\frac{LCS(X, Y)}{length(Y)}\), and \(P_{lcs}\) is precision, defined as \(\frac{LCS(X, Y)}{length(X)}\). \(LCS(X, Y)\) is the length of the longest common subsequence between the generated and reference texts, while \(length(X)\) and \(length(Y)\) are the sequence lengths. \(\beta\) is a weighting factor for recall.

While ROUGE is widely used due to its simplicity and alignment with human evaluations, it primarily focuses on lexical overlap rather than deeper semantic content.

BERTScore leverages a pre-trained BERT model to calculate semantic similarity between generated text and reference text:
$$
BERTScore = \frac{1}{|C|} \sum_{c \in C} \max_{r \in R} sim(c, r)
$$
where \(C\) is the set of candidate words, \(R\) is the set of reference words, and \(sim(c, r)\) is the semantic similarity between word pairs as determined by the BERT model.

\textbf{Confidence Evaluation}

Confidence evaluation assesses the reliability of LLM outputs to identify cases where the model is overly confident or underconfident, contributing to factual hallucinations.

ECE evaluates confidence calibration by measuring the alignment between a model's predicted confidence and its actual accuracy. The formula is:
$$
ECE = \sum_{m=1}^{M} \frac{|B_m|}{N} |acc(B_m) - conf(B_m)|
$$
where \(M\) is the number of probability intervals, \(B_m\) is a confidence bin, \(acc(B_m)\) is the accuracy for \(B_m\), and \(conf(B_m)\) is the average confidence for \(B_m\).

ECE can be influenced by the inherent correctness of events. To address this, MACROCE provides a more balanced confidence evaluation method by focusing on confidence in both correct and incorrect predictions \cite{si2022re}. Additionally, entropy-based methods offer an effective way to assess model uncertainty \cite{huang2024uncertainty}. Confidence monotonicity and uniformity of confidence intervals are also valuable for evaluating LLM calibration \cite{zhang2024calibrating, tomani2024uncertaintybasedabstentionllmsimproves}.

\textbf{Datasets}

When evaluating hallucination phenomena, the primary step is to precisely delineate the categories of hallucinations.
Datasets annotated with hallucination categories not only serve as the foundation for training detection systems\cite{gu2024anah,ridder2024hallurag} , but they can also be used for quantitatively evaluating the efficacy of these systems\cite{chen2024unified} .

Once hallucination types are accurately identified, targeted datasets can be constructed to specifically detect, evaluate, and mitigate various types of hallucinations.
For factual hallucinations, datasets can be built based on factual knowledge \cite{lin2021truthfulqa,elaraby2023halo} or through the contrast of true and false events \cite{azaria2023internal}.
Deeper explorations, such as unanswerable math problems proposed by Sun\textit{et al.}
\cite{sun2024benchmarking}, not only test factual hallucinations but also measure the model's awareness of the boundaries of knowledge \cite{amayuelas2023knowledge}, and promote its cognitive calibration \cite{yin2023large}.

For faithfulness hallucinations, the primary focus of detection lies in the depth of instruction comprehension and the extent to which the response adheres to the given instructions. DiSQ \cite{miao2024discursive} evaluates LLMs' understanding of event relationships, thereby assessing fidelity from a comprehension perspective. FaithDial \cite{dziri2022faithdial} and IFEval \cite{zhou2023instruction} examine the degree of compliance with instructions. The HaluQuestQA \cite{sachdeva2024fine} dataset provides a comprehensive evaluation of both aspects. Collectively, these datasets establish a systematic framework for assessing fidelity hallucinations, facilitating a deeper understanding of LLMs' potential biases and areas for improvement in instruction adherence and information fidelity.

Logical fallacy detection datasets are designed to rigorously assess whether large language models  maintain logical consistency in their responses—ensuring that all components remain coherent and directionally consistent in reasoning. Multi-hop reasoning tasks, which require extended thought processes, provide an ideal foundation for constructing such datasets \cite{geva2021did}. For instance, TriviaQA \cite{joshi2017triviaqa} requires LLMs to retrieve information across multiple sentences, whereas HotpotQA \cite{yang2018hotpotqa} involves retrieving information from multiple documents. Both datasets assess the models' ability to engage in reasoning, effectively testing their capacity to integrate knowledge and construct logically sound arguments.

Recent studies have further expanded the resources available for analyzing hallucination phenomena in LLMs. \cite{ridder2024halluragdatasetdetectingcloseddomain} introduced a dataset that leverages internal state representations—such as contextualized embedding vectors and intermediate activation values—to detect closed-domain hallucinations in Retrieval-Augmented Generation (RAG) systems. This resource, together with its publicly available code, provides a robust foundation for training classifiers capable of fine-grained hallucination detection. In parallel, \cite{chen2024hallucinationdetectionrobustlydiscerning} proposed a dataset specifically designed to differentiate between reliable and hallucinated outputs in LLM responses, offering a valuable benchmark for evaluating the factual consistency of generated answers. Moreover, \cite{gu2024anahv2scalinganalyticalhallucination} presented an automated framework based on an iterative self-training approach using the Expectation Maximization algorithm, which efficiently scales hallucination annotation datasets and significantly improves detection accuracy in zero-shot settings. Collectively, these contributions offer critical tools and methodologies for the systematic assessment and mitigation of hallucination phenomena in LLMs, thereby promoting the development of more reliable and trustworthy AI systems.

\textbf{Benchmark-Based Evaluation}

Benchmark-based evaluations are designed to focus on content from specific domains, where unverifiable content frequently leads to a high incidence of factual hallucinations. Common benchmarks include HaluEval by Li \textit{et al.} \cite{li2023halueval}, which targets general hallucination detection; Sun \textit{et al.}’s \cite{sun2024benchmarking} framework for evaluating unanswerable math problems; and Sachdeva \textit{et al.}’s \cite{sachdeva2024fine} evaluation for long-form question-answering. Rahman \textit{et al.} \cite{rahman2024defan} offer a comprehensive evaluation that addresses both factual and faithfulness hallucinations. Additionally, specialized benchmarks have been developed for vertical fields such as healthcare and law \cite{lin2021truthfulqa}. Chen \textit{et al.} \cite{chen2024unified} evaluate various approaches for mitigating hallucinations, while Su \textit{et al.} \cite{su2024unsupervised} propose a novel benchmark that leverages the internal states of LLMs to detect hallucinations.

Although these benchmarks are effective within their respective domains, they often lack broad applicability and pose risks of dataset leakage \cite{ni2024training, valentin2024costeffectivehallucinationdetectionllms}. Consequently, there is a pressing need to develop more universal and secure evaluation frameworks to ensure comprehensive and robust assessment across a wider range of contexts.

\subsubsection{Manual Evaluation}

Manual evaluation typically involves experts or crowdworkers who assess the quality of generated content based on subjective criteria. In scenarios requiring creative interpretation, cultural sensitivity, or emotional nuance, manual evaluation provides a depth of understanding that automated methods often struggle to achieve. Human evaluators are particularly skilled at detecting subtle shifts in tone, implied meanings, and the cultural appropriateness of content—areas where machines have difficulty making accurate judgments. From a practical perspective, manual evaluation can better reflect a model’s utility in real-world settings. By simulating realistic scenarios, this approach tests whether the model can consistently produce high-quality results in dynamic environments \cite{sachdeva2024fine}.

However, manual evaluation poses challenges, including high costs, time requirements, and the risk of subjective bias. Finding a sufficient number of evaluators with relevant expertise is both resource-intensive and time-consuming, and the results are often influenced by individual perspectives, potentially compromising objectivity and consistency. Additionally, human evaluators may struggle to identify certain technical flaws or statistical biases, which limits the effectiveness of manual evaluation in some areas \cite{burns2022discovering}.

\subsection{Detection of Hallucinations}

The presence of hallucinations not only undermines the reliability of LLMs but can also mislead users. As a result, effective detection and mitigation of hallucinations have become key areas of focus in LLM research. To further progress in this field, we examine recently proposed detection methods and categorize them to help improve the accuracy and trustworthiness of model outputs.

In practice, LLMs generally fall into two categories: white-box models, where users can access internal parameters, and black-box models, where users can only view the model’s output without insight into internal workings. This section provides a separate review of detection strategies tailored to each scenario.
\begin{figure}[t]
  \centering
  \includegraphics[width=0.95\columnwidth]{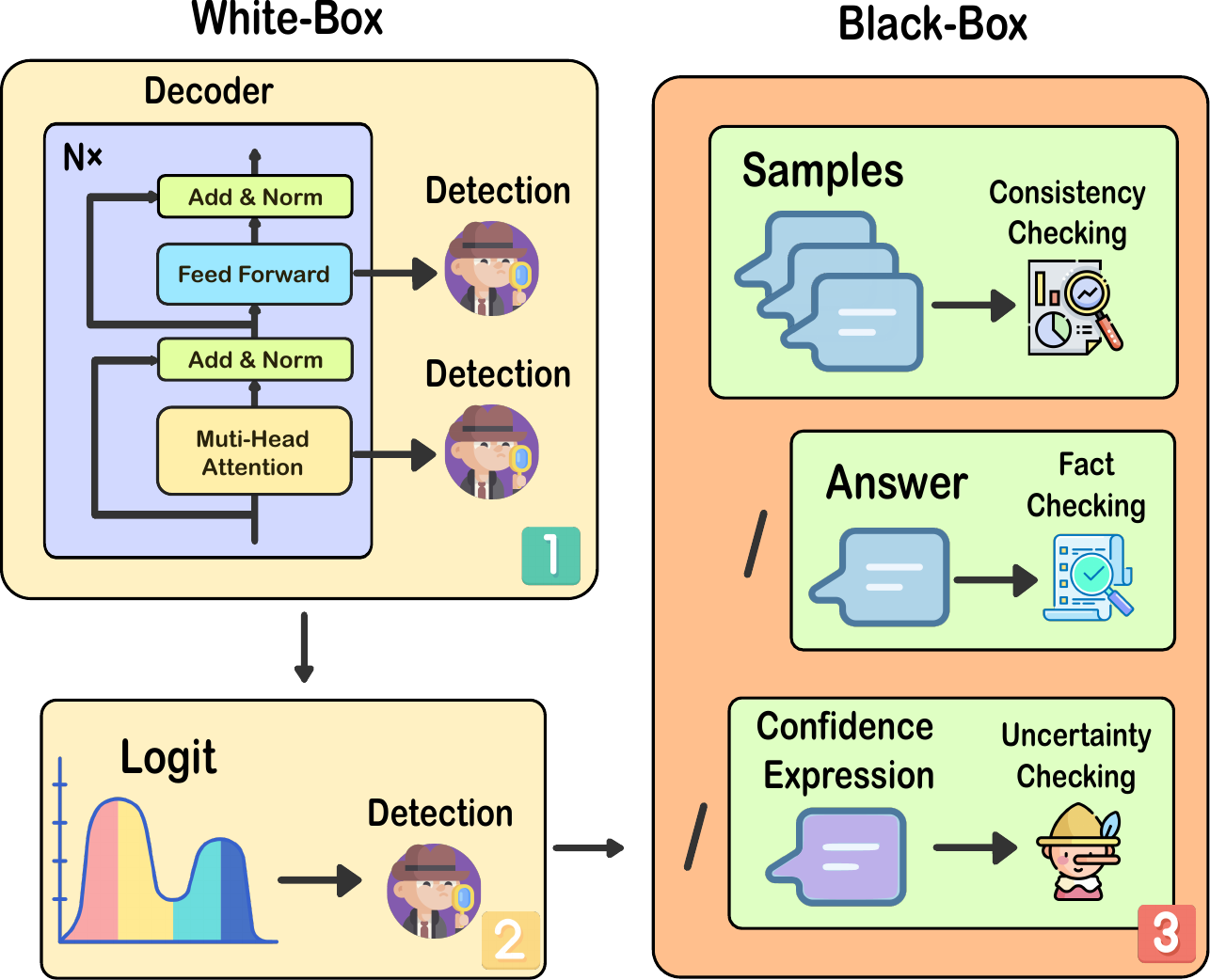} 
  \caption{\centering A Standard Paradigm of Detecting Hallucinations} 
  \label{fig:summary}
\end{figure}

\subsubsection{White-Box Model Detection}

For white-box models, which allow access to internal parameters of the LLM, mainstream research focuses on directly utilizing this information to detect hallucinations. When the model generates non-factual content, abnormalities in internal parameters can often be detected. For instance, Zhang \textit{et al.} \cite{zhang2024transferable} intermediate hidden layer representations are analyzed during response generation to assess the response's factual consistency. Similarly, Su \textit{et al.} \cite{su2024unsupervised} examines the context embeddings within the LLM and applies a multilayer perceptron (MLP) to classify these embeddings, distinguishing between hallucinatory and non-hallucinatory states.

In addition to embedding-based methods, Chen \textit{et al.} \cite{chen2024inside} explores neuron activation states. It was found that certain neurons in the penultimate layer exhibit abnormally high or low activation values, which can cause the model to produce overly confident but incorrect responses.

Moreover, the distribution of logits in the model’s output layer has been shown to contain cues about statement uncertainty. Ji \textit{et al.} \cite{ji2024llm} and Liu \textit{et al.} \cite{liu2024uncertainty} combine the logits from the LLM's final layer with neural networks to analyze and detect statement uncertainty effectively. Kumar \textit{et al.} \cite{kumar2024confidence} aligns each word’s internal token probability with the model’s self-assessed confidence, enhancing the LLM’s ability to recognize its hallucinations. Additionally, Kuhn \textit{et al.} \cite{kuhn2023semantic} and Farquhar \textit{et al.} \cite{farquhar2024detecting} improve upon traditional token-entropy estimations by generating multiple responses, sampling based on the model’s output distribution, and clustering similar responses, demonstrating that semantic entropy provides a more accurate measure of the model’s uncertainty.

Of particular note, Xiao \textit{et al.} \cite{xiao2021hallucination} categorizes uncertainty into epistemic (knowledge-based) and aleatoric (randomness-based) uncertainty. By analyzing logits and token probabilities, they found that epistemic uncertainty is often related to the model's knowledge limitations, while aleatoric uncertainty is tied to noise or randomness within the input data.

Recent advances have revealed additional pathways for addressing hallucination phenomena by tapping into alternative internal signals and novel decoding strategies. For example, one study identified “knowledge overshadowing” as a new mechanism leading to amalgamated hallucinations, and it proposed a decoding approach specifically designed to alleviate these effects \cite{zhang2024knowledgeovershadowingcausesamalgamated}. Similarly, the Lookback Lens framework exploits attention maps by analyzing contextual information and attention head outputs to classify hallucinations, subsequently introducing a context-based decoding strategy to mitigate such errors \cite{chuang-etal-2024-lookback}. Furthermore, another approach leverages graph structures to segment model outputs, thereby detecting hallucinations based on the structural relationships between tokens \cite{nonkes-etal-2024-leveraging}. Collectively, these methods enrich the white-box paradigm by integrating diverse internal representations and innovative decoding techniques, which together offer promising avenues for enhancing the factual consistency and reliability of large language models.

To summarize, detection methods for hallucinations in white-box models utilize access to LLM parameters, leveraging insights from embeddings, logits, and neuron activation states. By combining these internal indicators with external classifiers or neural networks, researchers can accurately identify potential hallucinations during content generation. Additionally, analyzing different uncertainty types—such as epistemic and aleatoric—provides a deeper understanding of how hallucinations form, offering theoretical and practical guidance for improving model accuracy and reliability.

\subsubsection{Black-Box Model Detection}

For black-box models, where the internal parameters of the LLM are inaccessible, detection becomes more challenging as analysis is limited to the model’s input and output. Unlike white-box methods, most research on black-box models aims to directly detect hallucinations based on the model’s outputs. We categorize black-box hallucination detection into three primary approaches: consistency-based analysis, confidence-based assessment, and the use of auxiliary models or tools.

Most existing detection methods focus on the first two categories. The first approach leverages consistency-based analysis to detect hallucinations. Recently, OpenAI introduced ChatGPT-o1, which uses the Chain of Thought (CoT) approach—a method that has shown significant promise in hallucination detection. For instance, Wu \textit{et al.} \cite{wu2024uncertainty} proposed a technique where the model independently answers a verification question and then uses CoT to answer it again based on the initial explanation. Comparing the consistency of these two answers helps identify potential hallucinations. A similar approach, Xue \textit{et al.} \cite{xue2023rcot} uses reverse reconstruction to generate the original question from the CoT-produced answer. Any inconsistencies between the reconstructed and original questions indicate hallucinations.

Another consistency-based method is multi-response analysis, as introduced by Manakul \textit{et al.} \cite{manakul2023selfcheckgpt}. The authors found that when a model generates hallucinations, its responses often show inconsistencies. By asking the same question multiple times and analyzing the consistency among the generated responses, this approach effectively detects hallucinations and pioneers the use of multi-response analysis.

In recent developments, Qiu \textit{et al.} \cite{qiu2023detecting} proposed a novel method by translating English text into multiple languages and adapting English consistency metrics to a multilingual context for consistency detection. Building on diverse response generation, Zhang \textit{et al.} \cite{zhang2023sac} introduces controlled perturbations into prompts to generate varied but semantically equivalent questions, enabling hallucination detection through response consistency.

Another effective method involves consistency fine-tuning, where models are trained to produce more consistent outputs. Park \textit{et al.} \cite{park2024mitigating} generates challenging yet natural dialogue prompts to induce hallucinations, which are then used to fine-tune the model, reducing the likelihood of hallucinations in dialogue tasks, especially those combining vision and language.

The second major category of black-box hallucination detection is based on the model’s confidence levels. An intuitive approach is to have the model express its uncertainty in natural language. Lin \textit{et al.} \cite{lin2022teaching}, Xiong \textit{et al.} \cite{xiong2023can} and Kim \textit{et al.} \cite{kim2024m} demonstrate that this approach is effective. For example, Yona \textit{et al.} \cite{yona2024can} introduces prompts like “answer directly” and “express uncertainty” to guide the model in producing answers in different styles, allowing researchers to assess the model’s ability to express uncertainty and detect potential hallucinations. Zhou \textit{et al.} \cite{zhou2023navigating} further found that when LLMs include phrases like “I think it is...,” performance improves, whereas overly confident expressions like “I am certain it is...” reduce accuracy.

However, eliciting uncertainty through natural language is not always reliable. Xiong \textit{et al.} \cite{xiong2023can} found that relying solely on the model’s verbal expressions of uncertainty may result in overconfidence or internal contradictions. The study showed that LLMs can overestimate the accuracy of their responses, even when their internal reasoning indicates high uncertainty.

The third approach to hallucination detection is introducing auxiliary models or tools. Pelrine \textit{et al.} \cite{pelrine2023towards} demonstrates that using a more powerful LLM to construct a confidence detector with prompts and assign scores can effectively identify hallucinations. Another study, Kirstein \textit{et al.} \cite{kirstein2024s} leverages CoT to enable an external GPT-4 model to function as a checker, examining whether summaries generated by the base model contain hallucinations. Instead of external models serving solely as checkers, Chang \textit{et al.} \cite{chang2024uncovering} encourages multiple LLMs to discuss the same input, generating diverse viewpoints to uncover biases or errors within the model.

Recent investigations have further enriched the black‐box hallucination detection landscape by introducing innovative analytical perspectives and decoding strategies. For instance, one study examines layer‐wise information deficiency by analyzing the representations associated with unanswerable questions and ambiguous prompts, revealing that a lack of sufficient informational cues at certain layers correlates with the emergence of hallucinations \cite{kim2024detectingllmhallucinationlayerwise}. In another approach, a framework termed HalluCana employs a “canary lookahead” mechanism during decoding to proactively flag and mitigate potential hallucinations, thereby enhancing output reliability \cite{li2024hallucanafixingllmhallucination}. Furthermore, the CALM framework leverages a curiosity‐driven auditing strategy based on reinforcement learning to both trigger and assess hallucination levels dynamically, offering a novel means of monitoring model performance without incurring the high computational cost of invoking a larger model \cite{zheng2025calmcuriositydrivenauditinglarge}. Collectively, these methods complement existing consistency‐ and confidence‐based approaches, providing additional pathways to robustly detect hallucinations in black‐box settings.

However, introducing another LLM for detection can slow down the process. Hu \textit{et al.} \cite{hu2024slm} addresses this by using a small language model (SLM) for initial hallucination detection. When the SLM identifies a potential hallucination, the LLM performs a deeper analysis, providing detailed explanations for inconsistencies, incoherences, or factual inaccuracies.

In summary, various effective strategies have been developed for detecting hallucinations in black-box models, focusing primarily on consistency analysis, confidence assessment, and using auxiliary models or tools\cite{deng2024llmgoodpathplanner}. Consistency-based methods compare multiple generated outputs to identify discrepancies, confidence-based methods assess the model’s uncertainty to detect errors, and the use of external models brings additional computational power and evaluation depth. While each approach has its strengths and limitations, together they significantly advance the detection of hallucinations in black-box models, contributing to the overall reliability of large language models.

\subsection{Methods for Mitigating Hallucinations}

This section explores advanced strategies aimed at mitigating hallucinations in LLMs. It delves into three main approaches: Shifting Demand Strategy, Task Simplification Strategy and Capability Enhancement Strategy.

\subsubsection{Shifting Demand Strategy}

Traditional methods typically focus on guiding LLMs toward generating accurate content without accounting for the varying difficulty of questions. 
As a result, these approaches may struggle when confronted with excessively complex queries. 
An alternative perspective is introduced by the Demand Shifting Strategy, which allows the LLMs to refrain from answering when there is a high likelihood of generating hallucinations. 
This strategy reduces the occurrence of hallucinations, with the central challenge being how to enable the LLMs to effectively assess when a query falls beyond its knowledge scope.
\begin{figure}[t]
  \centering
  \includegraphics[width=0.7\columnwidth]{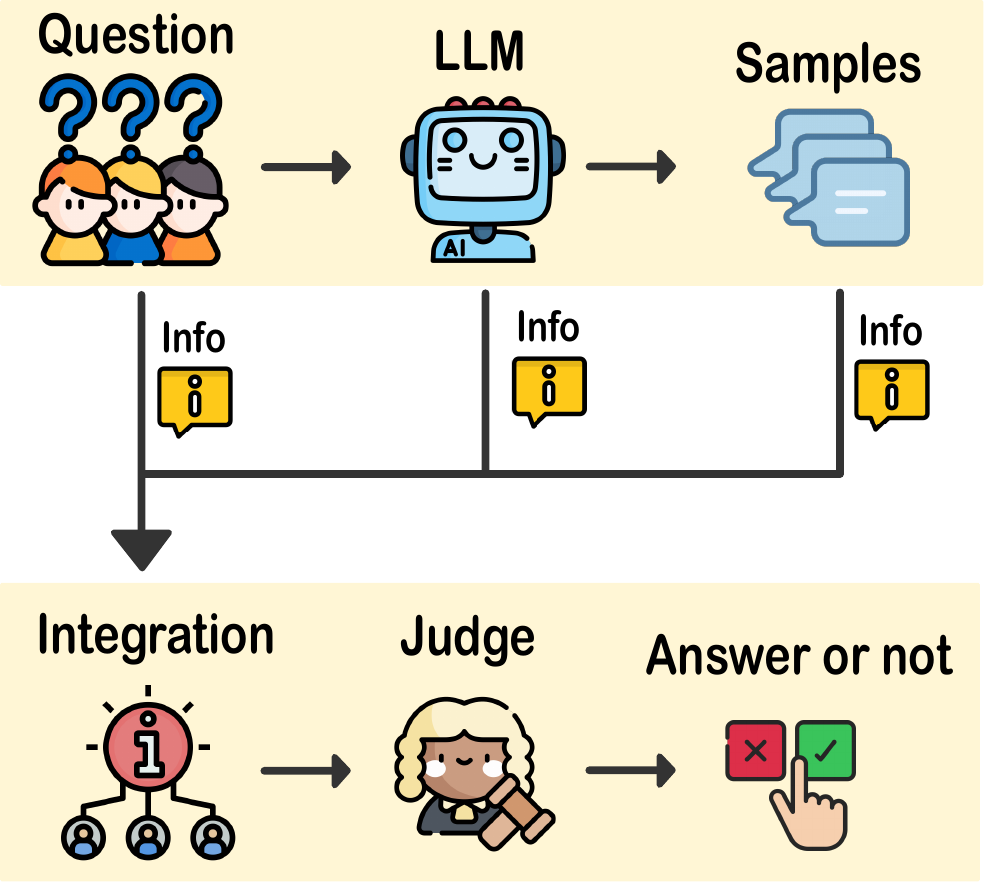} 
  \caption{\centering Paradigms of Tast Simplification Strategy} 
  \label{fig:summary}
\end{figure}

\textbf{Response Abstention and Confidence Calibration}

Given that LLMs cannot guarantee accurate responses to all queries, an effective strategy for mitigating factual hallucinations is to equip them with the ability to recognize their own uncertainty and abstain from responding when confidence is insufficient.
This approach entails enabling the LLM to assess its own reliability and opt not to respond when confronted with queries that fall beyond its knowledge boundary or those it fails to adequately comprehend.
However, current LLMs struggle to accurately discern these knowledge boundaries and frequently exhibit overconfidence in their responses \cite{ren2023investigating}.

Directly fine-tuning the LLM to selectively abstain from answering certain questions can enhance its ability to recognize and delineate the boundaries of its own knowledge \cite{amayuelas2023knowledge, zhang2024r}.
However, this approach may lead to excessive conservatism, prompting the model to abstain from answering queries that it is otherwise capable of handling accurately \cite{cheng2024can}. Therefore, achieving an optimal balance in the model’s confidence through rigorous confidence calibration is imperative, ensuring that it neither exhibits exorbitant confidence nor an excessive reluctance to respond.

Confidence calibration methods are generally classified into those integrated during the training phase and those deployed at the inference stage \cite{jiang2021can}.

During the training phase, fine-tuning and reinforcement learning (RL) methods, such as Proximal Policy Optimization (PPO), are frequently employed to facilitate the calibration of confidence levels in LLMs \cite{band2024linguistic, xu2024sayself, wu2024alleviating}. Additionally, as outlined in Section 3.4.2, LLMs generally exhibit robust confidence in their calibrated outputs following pre-training. Consequently, allowing post-training LLMs to re-acquire pre-trained features presents a promising approach to further enhancing calibration accuracy \cite{he2023preserving}.

An intuitive yet fundamental approach to confidence calibration is to explicitly enable LLMs to articulate their confidence levels within their outputs during the inference phase \cite{tian2023just}. However, due to potential discrepancies or biases in the training data, these self-reported confidence estimations may not always be reliable \cite{zhou2023navigating}.
Few-shot re-calibration represents a straightforward yet effective strategy for enhancing confidence calibration \cite{li2024few}. Additionally, approaches such as retrieval-augmented generation (RAG) \cite{ren2023investigating} and the incorporation of enriched contextual information \cite{yin2023large} have also demonstrated efficacy in improving calibration.

In addition to these overarching approaches, fine-grained modifications to the internal architecture of the model can offer deeper insights into the delineation of its knowledge boundaries. This aspect will be explored in detail in Section 4.3.3 on structural adjustments.

Through the implementation of effective confidence calibration, LLMs can more accurately abstain from responding to questions they are uncertain about \cite{yang2023improving}, thereby mitigating both overconfidence and undue conservatism in their outputs.

\subsubsection{Task Simplification Strategy}

Task Simplification Strategy aims to simplify the challenges LLMs face by using techniques such as Retrieval-Augmented Generation (RAG), Knowledge Graphics (KG) or prompt engineering, as well as by encouraging LLM to self-reflect and generate correct responses through multiple attempts, ultimately reducing the probability of hallucinations.

\begin{figure*}[t]
  \centering
  \includegraphics[width=\textwidth]{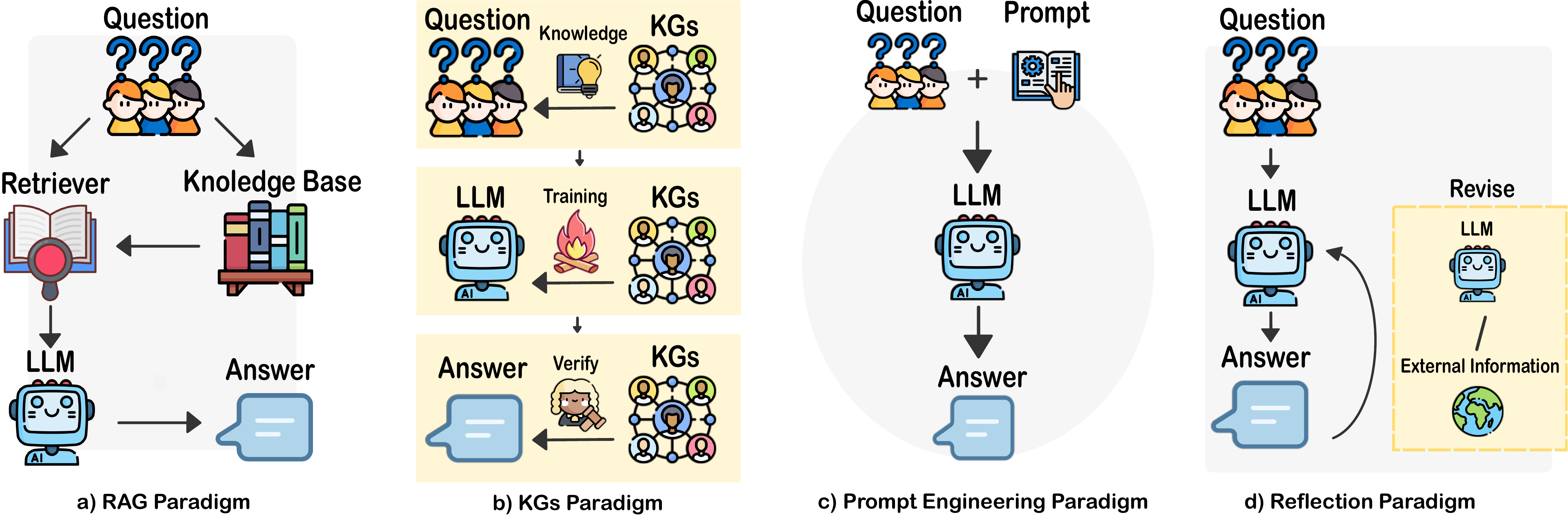} 
  \caption{\centering A Standard Paradigm of Shifting Demand Strategy} 
  \label{fig:summary}
\end{figure*}

\textbf{Retrieval-Augmented Generation}

Given the intrinsic limitations of LLMs, including potential gaps in their knowledge base and susceptibility to outdated training data, the integration of external information is a widely adopted strategy for enhancing response accuracy and recency.
Retrieval-Augmented Generation (RAG) addresses these challenges by retrieving relevant external knowledge, which the LLM can then leverage to enhance response fidelity and mitigate factual hallucinations. The effective implementation of RAG necessitates two fundamental considerations: the selection of the most relevant knowledge for retrieval and the seamless integration of this information into the response generation process.

From a retrieval perspective, leveraging a diverse range of information sources is essential for improving coverage and reliability \cite{li2023chain}. Incorporating heterogeneous databases enhances the breadth of knowledge available to the model, facilitating the generation of more nuanced and comprehensive responses. Furthermore, structuring and partitioning the knowledge base can refine the retrieval process by ensuring that only the most pertinent content is selected \cite{wang2024m}. Prioritizing the retrieval of ``long-tail'' knowledge—specific or less frequently encountered information that is often absent from mainstream sources—can further augment model performance and retrieval efficiency \cite{li2024role}. This targeted retrieval approach is particularly beneficial in addressing knowledge gaps inherent to LLMs. Beyond traditional database queries, specialized retrieval mechanisms—such as smaller, fine-tuned models designed for real-time updates or domain-specific knowledge extraction—offer valuable supplementary resources for LLMs \cite{feng2023knowledge}. These domain-adaptive retrieval systems provide timely and contextually relevant insights that may otherwise be inaccessible. Once relevant knowledge is retrieved, additional filtering mechanisms can be employed to distill key insights while eliminating extraneous details or noise that might obscure the model’s comprehension \cite{anjum2024halo, zhu2024information}.

During the generation phase, the LLM is expected to adjust its strategy to optimize the output, contingent on the completeness and quality of the external information retrieved.
Should the retrieved content be comprehensive and adequately address the query, the LLM can function as a summarizer, consolidating and presenting the information with clarity. However, in cases where the retrieved content is incomplete or insufficient, the LLM must integrate its internal parametric knowledge—pre-existing information embedded within its model parameters—in order to generate a coherent and accurate response \cite{xu2024unsupervised}.
Challenges may emerge when the retrieved texts are overly lengthy or exhibit complex structures, as such factors can impede the LLM's ability to efficiently process and utilize the content. Advanced attention mechanisms, however, can assist the model in concentrating on the most pertinent aspects of the retrieved data, thereby enhancing its processing capabilities \cite{yoon2024listt5}.
Moreover, maintaining contextual coherence and preventing its fragmentation across segmented inputs is critical for optimizing the LLM's comprehension and the fidelity of its generated responses \cite{qian2024grounding}.
Notably, even when retrieved information is factually accurate, ensuring that the generated responses maintain both factual integrity and contextual coherence remains a persistent challenge\cite{niu2023ragtruth, magesh2405hallucination}.`` 
Sun \textit{et al.} identified that LLMs tend to prioritize internal parametric knowledge over external information, as evidenced by the dominance of feed-forward networks (FFNs) in the residual stream and the inefficiency of copying heads in integrating retrieved data. They addressed this by balancing these capabilities, effectively mitigating the issue of incomplete knowledge integration \cite{sun2024redeep}.

In conclusion, RAG constitutes a fundamental approach to augmenting the capabilities of LLMs through the integration of external knowledge sources. However, to fully harness its potential, critical advancements are required in the domains of retrieval precision, computational efficiency, knowledge synthesis, and contextual alignment.
Future research efforts should prioritize the refinement of these critical dimensions, with a particular focus on the optimization of retrieval methodologies, the enhancement of contextual integration frameworks, and the improvement of alignment mechanisms between the model's parametric knowledge base and external retrieval systems. Such advancements are imperative to mitigate existing limitations and to further augment the efficacy and performance of RAG-enhanced language models.

\textbf{Knowledge Graph} 

A Knowledge Graph (KG) is a structured semantic database that organizes entities and their relationships in the form of a graph, providing a powerful framework for representing complex, interconnected information \cite{hogan2021knowledge}.
By using this structured format, KGs facilitate the extraction and organization of knowledge, which can significantly enhance the performance of LLMs. \cite{agrawal2023can}. Specifically, KGs have been applied in detecting hallucinations generated by LLMs \cite{guan2024mitigating} and in offering effective strategies to mitigate them \cite{panda2024holmes}.
Serving as both a source of knowledge and a tool for evaluating accuracy, KGs play a critical role in improving the coherence and factual correctness of LLM-generated content.

However, leveraging Knowledge Graphs (KGs) to mitigate hallucinations introduces challenges akin to those encountered in RAG. Key challenges include:
(1) determining which specific graph information should be retrieved based on the given context or query, and
(2) ensuring that the retrieved information is seamlessly integrated to enhance the quality and factual accuracy of the generated response.

In the retrieval phase, capturing the often incomplete and nuanced relationships between entities within the KG is of paramount importance. Attention mechanisms have proven particularly effective in this regard, facilitating the retrieval of meaningful relationships between nodes within the graph \cite{liu2024knowformer}.
These mechanisms help ensure that only relevant knowledge is retrieved, thereby enhancing the contextual understanding and coherence of the LLM’s outputs. Furthermore, contrastive learning techniques have demonstrated their ability to mitigate noise and inconsistencies present within both the graph structures and the textual data, thereby improving the quality of the retrieved information \cite{jiang2022unikgqa, liu2024knowledge}.
Through the refinement of the retrieval process, these techniques ensure that the data fed into the LLM is both accurate and pertinent, thereby enhancing the generation of coherent, contextually relevant, and factually precise outputs.

Following the retrieval process, Knowledge Graphs (KGs) offer substantial advantages during both the training and inference phases.
During training, the integration of KGs enables LLMs to acquire a more comprehensive understanding of graph structures and inter-entity relationships, which enhances the model's capacity for knowledge representation and supports complex reasoning tasks—particularly those that require multi-hop inference across diverse domains \cite{gao2024two}.
This foundational understanding equips the model to better process and generate responses to intricate queries.
In the inference phase, KGs can be effectively leveraged by converting user inputs into structured graph queries, allowing for the targeted retrieval of relevant information from graph databases.
The retrieved knowledge is then integrated into the model’s output generation process, thereby improving the factual accuracy and informational depth of the response, while ensuring the generated content is contextually relevant and factually sound \cite{pusch2024combining}.

Additionally, KGs exhibit considerable potential when integrated with other techniques to mitigate hallucinations.
The incorporation of RAG  enhances the retrieval process by introducing structured relationships that significantly elevate the quality and relevance of retrieved information. 
Unlike conventional RAG frameworks, which often result in redundant or fragmented data retrieval, knowledge graphs (KGs) provide structured, interlinked factual knowledge that effectively mitigates these shortcomings. This structured representation enables LLMs to produce more coherent, precise, and contextually nuanced responses, thereby enhancing the overall quality of generated outputs.
Similarly, the integration of KGs with Chain of Thought (CoT) techniques further enhances LLM performance by embedding structured knowledge and facilitating iterative, step-by-step reasoning, thereby effectively addressing factual hallucinations, logical inconsistencies, and improving overall response accuracy and coherence \cite{sanmartin2024kg}.

Finally, knowledge graphs (KGs) serve as an effective tool for evaluating the quality of large language model (LLM) outputs. Through systematic cross-referencing of generated content with the structured and validated knowledge embedded within KGs, potential hallucinations undergo systematic detection and methodical correction, ensuring greater factual accuracy and reliability in generated outputs \cite{sansford2024grapheval}. This methodology facilitates a more rigorous and systematic assessment of LLM performance, supporting iterative fine-tuning processes that enhance the generation of high-quality, factually reliable, and contextually coherent outputs.

In conclusion, effectively leveraging KGs across training, inference, and evaluation stages allows LLMs to improve the quality of their outputs significantly.
These improvements are particularly noticeable in enhanced factual accuracy, logical consistency, and reduced hallucinations, leading to more trustworthy and contextually appropriate language generation.
Future research efforts should prioritize the enhancement of scalability, knowledge accuracy, and integration mechanisms to address existing limitations and further optimize the performance of LLMs.

\textbf{Prompt Engineering}

Prompt engineering represents a systematic methodology for optimizing LLMs generative capabilities through strategically designed input stimuli, characterized by its intrinsic capacity to simplify complex tasks via templated natural language instructions that decompose problems into modular subtasks, eliminate parametric training overhead through non-parametric implementation, and dynamically reconfigure semantic-syntactic templates to accommodate cross-task requirements. 
This section focuses on the evolutionary trajectory of the Chain-of-Thought (CoT) paradigm and its derivative optimization strategies, specifically analyzing their operational frameworks and engineering implementation schemas within the context of computational efficiency and task adaptability.

The Chain-of-Thought (CoT) paradigm enhances large language models' (LLMs) reasoning fidelity by structuring prompts to emulate human-like multi-step cognitive processes \cite{wei2022chain}.
This approach enforces a systematic decomposition of complex problems into sequential reasoning phases, beginning with global contextual framing and progressing through iterative refinement of intermediate conclusions. 
Such structured cognition mitigates the model's propensity for heuristic shortcuts during autoregressive decoding, thereby improving output coherence through explicit constraint satisfaction.
Building upon this foundation, advanced methodologies like least-to-most prompting \cite{zhou2022least} implement dynamic programming-inspired decomposition, where each reasoning step generates verifiable subgoals that recursively constrain subsequent inferences. 
This dual mechanism ensures both forward consistency (logical alignment across steps) and backward traceability (reducibility to primitive reasoning units), particularly effective in combinatorial optimization scenarios.

Modern implementations further demonstrate LLMs' capacity for meta-cognitive adaptation, wherein models autonomously generate task-specific CoT templates through in-context learning \cite{zhang2022automatic} and dynamically adjust reasoning trajectories via real-time analysis of epistemic states \cite{press2022measuring}.
The CoT ecosystem has subsequently evolved through specialized architectural variants: Auto-Critique frameworks integrate self-assessment loops that detect logical inconsistencies via counterfactual simulation \cite{mundler2023self}, while IAO prompting \cite{diallo2025iaopromptingmakingknowledge} enforces bidirectional verification between parametric knowledge and external evidence through structured checkpoints. Complementing these, the HalluciBot (H4R) system \cite{watson2024thingbadquestionh4r} employs probabilistic risk estimation via Monte Carlo sampling over perturbed query ensembles, enabling preemptive hallucination mitigation through adaptive query rewriting and compute-aware routing.

Collectively, these advancements transform CoT from a basic prompting technique into a robust framework for computational epistemology, where formalized reasoning protocols interact synergistically with the model's parametric knowledge. 
The resultant paradigm not only addresses hallucination through multi-layered verification but also maintains implementation efficiency via non-parametric adaptation, establishing new standards for reliable knowledge-intensive generation.

\textbf{Reflection}

Reflection in large language models (LLMs) can be conceptually partitioned into two distinct categories: self-reflection and environmental reflection. Self-reflection pertains to the model’s intrinsic capability to critically evaluate and identify its own errors, whereas environmental reflection harnesses external resources—such as alternative models, compiler error messages, or retrieved external knowledge—to detect and rectify mistakes. 

Regarding self-reflection, LLMs can systematically decompose their responses into discrete segments to rigorously assess the factual accuracy of each component \cite{zhang2024self}. Furthermore, by generating multiple responses from distinct perspectives and conducting comparative analysis, the model can evaluate internal consistency and logical coherence across its outputs \cite{zhang2024selfcontrastbetterreflectioninconsistent}. The incorporation of confidence calibration mechanisms further refines this process by enabling probabilistic estimation of potential errors based on the model’s confidence distribution across individual response elements \cite{chen2023adaptation}.

In contrast, environmental reflection involves soliciting feedback from external sources. One effective approach entails orchestrating a debate between two models on an identical query, which fosters the generation of a more comprehensive and balanced response \cite{chang2024uncovering}. Furthermore, the incorporation of external knowledge and the application of contextual contradiction detection techniques serve to identify and correct both factual inaccuracies (i.e., hallucinations) and logical errors in the model’s output \cite{huang2024queryagent}.

In summary, self-reflection empowers LLMs to internally scrutinize their reasoning and accuracy, while environmental reflection employs external checks and balances to further enhance response quality. The synergistic application of these reflective strategies is instrumental in mitigating hallucinations and logical inconsistencies, thereby augmenting the overall reliability and accuracy of LLM-generated outputs.

\subsubsection{Capability Enhancement Strategy}

Capability Enhancement Strategy involves enhancing a model’s performance on specific tasks through fine-tuning (FT) and structural optimization, which in turn helps to reduce hallucinations.

\textbf{Fine-tuning}

Fine-tuning LLMs is one of the most straightforward ways to tackle hallucinations in specific domains. Although naive fine-tuning to introduce new knowledge may linearly increase hallucinations (as discussed in 3.4.2), carefully crafted fine-tuning strategies can effectively mitigate different types of hallucinations. For instance, ``Learning Confidence for Transformer-based Neural Machine Translation'' focuses on adjusting the LLM’s confidence to ensure better calibration, reducing overly confident yet inaccurate outputs. Moreover, dynamically modifying the loss function based on task complexity can better align the model's training with its pre-trained features \cite{wang2024uncertainty}. Additionally, differentiating training objectives according to the nature of the task can help manage underconfidence caused by misalignment issues \cite{lin2024flame}.

However, FT approaches often suffer from high computational costs and long training times \cite{mitchell2022a}.This limitation has led to the emergence of more lightweight and efficient fine-tuning methods which are especially well-suited for addressing problems requiring up-to-date information \cite{mitchell2022a}. For example, prefix-tuning preserves the original parameters of the LLM while introducing trainable pseudo-prompts \cite{li2021prefix}, significantly reducing hallucinations across multiple tasks \cite{jones2023teaching}. Another example is the LoRA (Low-Rank Adaptation) approach, which keeps the original weights of the LLM frozen but introduces two trainable low-dimensional matrices to influence deep-layer neural responses, thus affecting the final output predictions \cite{yu2024interpreting}. AlphaEdit uses a null-space constrained knowledge editing method, where perturbations are projected onto the null space of preserved knowledge, ensuring that the output of post-edited LLMs remains consistent with previously learned information \cite{fang2024alphaedit}. Similarly, Zhao \textit{et al.} \cite{zhao2023knowledgeable} designs an efficient tuning network that integrates with a frozen-parameter LLM using adapters, which helps to prevent catastrophic forgetting during training. Additionally, pinpoint tuning of a limited number of attention heads can effectively mitigate hallucinations while minimizing knowledge loss \cite{chen2024yes}.These methods enable the efficient introduction of new knowledge while preserving the model's original performance, making it a more economical and scalable alternative to FT.

There have been notable advances in understanding the underlying mechanisms of hallucinations. Researchers are investigating subspaces within LLMs that represent ``universal truthfulness'' in order to address factual hallucinations \cite{liu2024universal}. However, generalization remains a significant challenge \cite{orgad2024llms}. By comparing consistency within activation spaces, unsupervised learning techniques improve the LLM's ability to capture and represent features accurately, thereby enabling the detection of errors that may otherwise be overlooked by humans \cite{burns2022discovering}. These developments highlight the limitations of current hallucination evaluation frameworks and suggest new avenues for further improvement.

In summary, Capability Enhancement Strategy through targeted fine-tuning and structural adjustments allows LLMs to improve their performance on specific tasks while minimizing hallucinations. These approaches hold promise for developing more accurate and reliable language models.
\begin{figure*}[t]
  \centering
  \includegraphics[width=\textwidth]{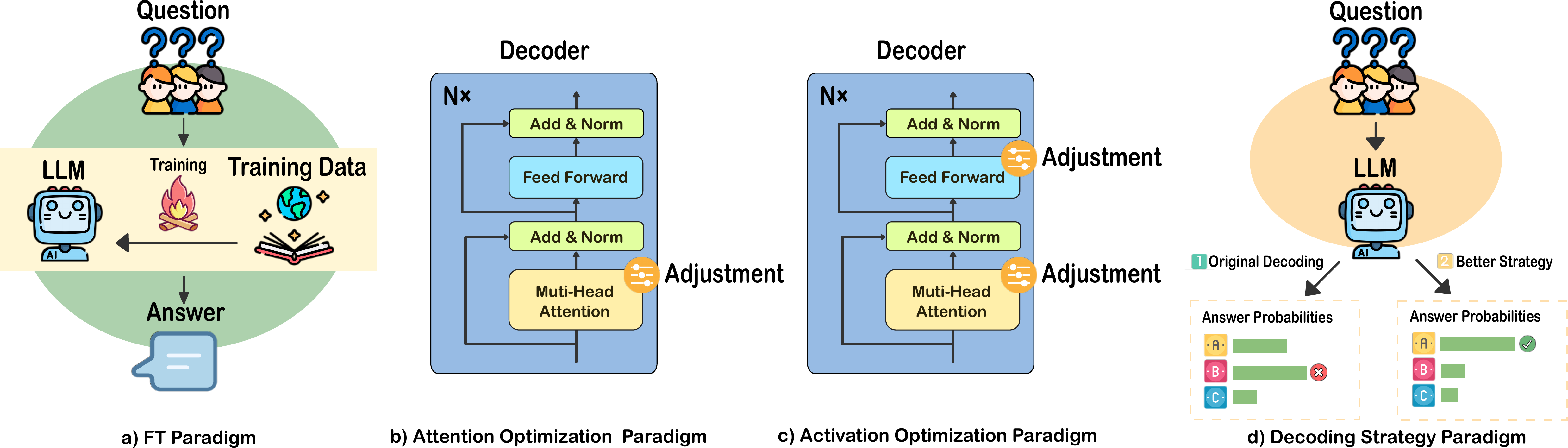} 
  \caption{\centering A Standard Paradigm of Shifting Demand Strategy} 
  \label{fig:summary}
\end{figure*}

\textbf{Structural Optimization}

LLMs have inherent structural constraints that are difficult to overcome, which makes optimizing their attention mechanisms, activation functions, and decoding strategies crucial for improving performance. This section will briefly introduce the generation mechanism of LLMs and explore various structural adjustment methods to mitigate hallucinations.

Investigating model structures is critical, as the internal states of LLMs play a vital role in estimating model uncertainty \cite{liu2024uncertainty, mehta-etal-2024-halu} and delineating the boundaries of their knowledge \cite{liang2024internal}. This understanding can assist in confidence calibration and help reduce factual hallucinations. Furthermore, refining attention mechanisms and decoding strategies can effectively address faithfulness hallucinations and logical inconsistencies.

The generation process of LLMs typically involves an embedding layer, multiple transformer layers, a linear transformation layer, and a final softmax layer. In simple terms, for an input sequence $\mathcal{T} = \{x_t\}_{t=1}^{T}$, the embedding layer $H_0$ converts discrete tokens into continuous vector representations $E[\mathcal{T}]$, setting the stage for further processing by subsequent transformer layers. In each $l$-th layer, the transformation layer $H_l$ performs a multi-step process: it multiplies the input with key-value matrices $P_l^h$, concatenates the resulting vectors to form multi-head attention, and then applies a weight matrix $W_l$. The transformed output passes through two stages of add \& normalize (A\&N) operations and a feed-forward network (FFN), which ensures proper layer normalization and transformation before feeding into the next layer. This iterative process continues until the final layer. Finally, the output is linearly transformed and passed through a softmax function to generate a probability distribution over the next possible token $P(x_{T+1}|x_{<T+1})$. The process is described by the following equations:
\begin{itemize}
    \item The input sequence and its embedding:
    $$
    \mathcal{T} = \{x_t\}_{t=1}^{T}, \quad H_0 = E[\mathcal{T}]
    $$

    \item Transformation at each layer with attention and normalization:
    \begin{equation*}
        H_{l+1} = A \& N\left(FFN\left(A\& N\left(W_l \cdot \text{Concat}_{h=1}^{H} (P_l^h \cdot H_l)\right)\right)\right)
    \end{equation*}

    \item Output probability distribution for generating the next token:
    $$
    P(x_{T+1}|\mathbf{x}_{<T+1}) = \text{Softmax}(\text{Linear}(H_N))
    $$
\end{itemize}

Through structural optimization, focusing on enhancing these layers and their operations, LLMs can achieve better performance and reliability. Improvements in attention mechanisms, activation functions, and decoding strategies contribute not only to reducing hallucinations but also to producing more coherent and contextually accurate outputs.

\textbf{Attention Mechanism}

Optimizing the attention mechanism is crucial for enhancing both the memory retention and reasoning abilities of large language models (LLMs). The attention mechanism allows the model to prioritize different parts of the input sequence, improving its ability to process and recall context during text generation. For example, enabling the model to focus on multiple passages simultaneously enhances memory and retrieval accuracy, ensuring that critical context is not overlooked \cite{yoon2024listt5}. This multi-focus capability improves the model's ability to synthesize information from different parts of a text, leading to more coherent and context-aware responses.

Additionally, certain attention heads are highly specialized and linked to specific tasks. For instance, arithmetic reasoning is handled primarily by a few specialized attention heads, highlighting their role in complex computational tasks \cite{yu2024interpreting}. This specialization suggests that fine-tuning or isolating these heads could boost the model’s performance in specific areas, increasing overall efficiency for those tasks.

LLMs can also be equipped with mechanisms that allow them to self-interpret the function of their attention heads, providing natural language descriptions of what each head is focusing on \cite{chen2024selfie}. This self-interpretation offers insights into the model’s internal processes, helping to understand how different layers and attention heads contribute to the final output. These descriptions can serve as probes for analyzing the intermediate layers of the model \cite{alain2016understanding}. Furthermore, techniques like path patching can be used to explore how information flows through the model, deepening our understanding of its interpretability and reasoning pathways \cite{wang2022interpretability}.

Since LLMs often emphasize information processed in both the early and final layers \cite{liang2024internal}, introducing adaptive structures in the final layer can help the model better utilize outputs from intermediate layers. This creates a more balanced distribution of attention across all layers, allowing the final layer to adaptively incorporate critical information that might otherwise be underemphasized \cite{verma2024adaptive}. Such layerwise adaptations not only enhance coherence and context in the generated responses but also improve the model’s ability to handle complex or long-form content more effectively.

\textbf{Activation Space}

Research on mitigating hallucinations by leveraging internal model information has advanced rapidly. One prominent direction involves representation editing methods, which explore the potential for effectively addressing hallucinations by utilizing information from the model’s activation space. Notably, identifying key attention heads and adjusting them based on activation space insights has emerged as a reliable paradigm \cite{li2024inference}. In addition to two intuitive improvements—enhancing detection accuracy \cite{chen2024truth} and refining the selection of activation space features \cite{zhang2024truthx}—improving generalization remains a critical objective \cite{orgad2024llms}. A more fine-grained mapping between task-specific capabilities and their corresponding representations in the activation space could offer greater flexibility in mitigating various types of hallucinations.

Another promising approach leverages internal model information to enhance other hallucination mitigation techniques. Specifically, assessing model confidence and its understanding of its own knowledge boundaries can aid in reducing hallucinations from a confidence calibration perspective. A practical recalibration strategy involves suppressing excessively confident layers to prevent any single layer from disproportionately influencing the decision-making process. This not only improves confidence calibration but also ensures a more balanced and stable response generation \cite{chen2024inside}. 

In summary, investigating the activation space is essential for understanding the internal mechanisms of LLMs, which in turn enhances our ability to control these models and mitigate hallucinations.

\textbf{Decoding Strategy}

Optimizing decoding strategies often involves refining the information generated by different layers and attention heads, improving the softmax process to enhance output quality, and addressing various forms of hallucinations. For example, faithfulness hallucinations and logical inconsistencies can be mitigated by using a context-aware output probability distribution $P_{\mathcal{T}|context}$ and subtracting a context-independent probability scaled appropriately, $P_{\mathcal{T}}$, to produce more contextually consistent results \cite{shi2023trusting}. Another effective strategy is to generate multiple outputs that consider different aspects of context and then select the most suitable one \cite{chuang2024lookback}. Suppressing attention to less relevant information also aids the LLM in focusing on the user's primary concerns \cite{zhang2024paying,zhao2024enhancing,gema2024decore}.

Furthermore, contrastive approaches using prompts \cite{wang2024mathbb} or specialized training methods \cite{zhang2023alleviating,yang2024improving} can help create models more prone to hallucinations, which, when compared with standard models, aids in reducing factual errors. Additionally, analyzing the consistency of thought between earlier and later layers provides insights that can effectively guide the decoding process to mitigate factual hallucinations \cite{chuang2023dola}.

In addition to optimizing decoding strategies for white-box models, innovative approaches for black-box LLMs have also gained attention. For instance, lightweight adapters that modify an LLM’s hidden states externally can generate improved responses without needing full access to internal parameters, offering a novel way to address hallucinations in black-box models \cite{cui2022decoder}.

By focusing on both activation space and decoding strategies, LLMs can be better calibrated, more contextually accurate, and more robust against various forms of hallucinations, enhancing their overall performance and reliability.

\section{Limitations and Future Outlook}

\subsection{Evaluation and Detection}

Current evaluation metrics for hallucinations face several significant limitations, including a lack of universal applicability, inadequate comprehensiveness, and the risk of dataset leakage\cite{yang2025hallucinationdetectionlargelanguage}. To evaluate hallucinations effectively, fairly, and holistically, there is an urgent need for an innovative evaluation system.

In practical applications, such as search engines and conversational AI, the manifestations of hallucinations often differ from those considered by existing benchmarks. Therefore, it is crucial to establish an automated detection framework tailored to real-world scenarios. Additionally, such a system must account for cost-effectiveness, performance optimization, and system stability to ensure practical and reliable solutions\cite{ravi2024lynxopensourcehallucination}.

A comprehensive evaluation framework should address multiple aspects of hallucinations, such as their frequency, type, severity, and impact on user experience. Existing methods often rely on simple metrics, like factual accuracy or coherence, which are insufficient to capture the full range of hallucination behaviors. For example, a hallucination in a medical dialogue system could result in dangerous misinformation, while a similar hallucination in casual conversation might lead only to a minor misunderstanding. Thus, evaluation must be context-sensitive, considering the specific domain and user expectations.

Balancing automated and manual evaluation approaches is a key challenge. Manual evaluations provide depth and contextual understanding but are resource-intensive, prone to subjective biases, and lack scalability for large datasets or continuous model assessments\cite{bavaresco2024llmsinsteadhumanjudges}. In contrast, automated evaluations offer efficiency and scalability but often struggle to detect nuanced errors or context-specific hallucinations. A hybrid evaluation system, leveraging human-in-the-loop mechanisms, could be a promising direction. Here, human oversight refines and guides automated metrics, ensuring both thoroughness and scalability.

One area of potential research in automated detection is the development of context-aware metrics that better align with human judgment. For instance, semantic similarity measures, such as BERTScore, provide a more nuanced assessment of coherence and relevance in generated content than traditional n-gram overlap metrics like BLEU or ROUGE\cite{chen2024hallucinationdetectionrobustlydiscerning}. However, these approaches still need further refinement to distinguish between minor stylistic differences and significant factual inconsistencies.

Another challenge is detecting context-specific hallucinations, which requires models to understand not only factual accuracy but also domain-specific norms and expectations. This is particularly difficult when evaluating models across diverse applications, as the criteria for hallucinations may vary greatly\cite{dahl2024large,zuo2025medhallbenchnewbenchmarkassessing}. One potential solution is to integrate retrieval-based augmentation during evaluation, where models are tested based on their ability to accurately retrieve and incorporate external knowledge. This could bridge the gap between general-purpose benchmarks and domain-specific expectations, providing a more holistic evaluation of a model's adherence to known facts.

Dataset leakage in evaluation benchmarks remains a persistent issue. If test data overlaps with the model's training data, evaluation results may be inflated, giving a false sense of the model's capabilities in detecting and mitigating hallucinations\cite{li-etal-2024-open-source}. Techniques such as data deduplication and the creation of dynamic, continuously updated test sets can mitigate this risk, ensuring that evaluation metrics reflect the model's true performance on unseen data\cite{huang2025thinkbenchdynamicoutofdistributionevaluation}.

Cost-effectiveness is another critical factor in developing an evaluation and detection framework. Evaluating hallucinations in real-time applications, like conversational agents or translation systems, requires low-latency, lightweight methods that do not overly burden system performance. This may call for the development of specialized evaluation tools optimized for different deployment environments, ensuring that hallucination detection remains efficient and scalable\cite{su2024unsupervisedrealtimehallucinationdetection}.

The performance of detection frameworks must also be robust across model updates and iterations. LLMs frequently undergo updates to improve performance or expand their knowledge base, so an effective evaluation system must adapt without requiring constant recalibration\cite{wang2024benchmarkselfevolvingmultiagentframework}. This requires developing generalizable metrics and detection strategies that remain valid across model variations, allowing continuous monitoring of hallucinations even as models evolve \cite{zhu2024dynamicevaluationlargelanguage}.

Finally, user-centered feedback is essential in hallucination evaluation. While technical metrics are vital for model development, user perception and satisfaction ultimately determine a model's success in real-world applications. Gathering user feedback on perceived hallucinations—including their type, impact, and frequency—can offer valuable insights into a model's real-world performance. Incorporating this feedback into the evaluation loop helps align model outputs with user expectations and provides data-driven guidance for further fine-tuning\cite{awsLLM2023}.

In conclusion, an effective evaluation and detection system for LLM hallucinations must be multi-dimensional, incorporating metrics that capture factual accuracy, coherence, context relevance, and user satisfaction. It should balance automation with human oversight, be tailored to domain-specific needs, and ensure cost-effectiveness without sacrificing performance or scalability. Such a comprehensive system is crucial for understanding and mitigating hallucinations in LLMs, leading to models that are both reliable and practical for widespread deployment.

\subsection{Capability Exploration}

The emerging capabilities of large-scale models are heavily influenced by scaling laws, such as the Chinchilla/Hoffman and Kaplan Scaling Laws, which highlight the complex relationships between model performance, parameter count, and data volume\cite{10.5555/3716662.3716692}. However, relying solely on increasing parameters and expanding datasets has reached its practical limits, making the enhancement of data quality a key focus.

\subsubsection{Enhancing Data Quality and Training Strategies}

The quality and diversity of training data significantly affect LLM performance\cite{pang2024improvingdataefficiencycurating}. As models scale, merely adding more data is insufficient unless that data is highly relevant and representative of the tasks the model is expected to handle. Ensuring datasets cover a wide range of linguistic styles, cultural nuances, and domain-specific knowledge is crucial for improving the model’s generalizability. However, curating high-quality data is a challenge, as noisy, biased, or overly narrow information can lead to poor generalization and even reinforce undesirable behaviors in the model’s outputs.

During training, one primary goal is to reduce the need for extensive post-training alignment \cite{zhou2024lima}. This involves integrating alignment objectives earlier in pre-training to ensure efficient and accurate model training\cite{liang2024alignment}. Pre-training tasks should better reflect end-use scenarios, focusing on truthfulness, factual recall, and logical consistency. However, care must be taken when fine-tuning for task-specific performance, as overfitting to certain alignment tasks may negatively affect the model's broader capabilities and increase the risk of hallucinations or confidence calibration issues.

Exploring transfer learning and domain adaptation offers a promising direction. Through transfer learning, LLMs can be adapted to specific domains using relatively small amounts of high-quality, targeted data, reducing the need for retraining on vast datasets\cite{NEURIPS2024_ea3f85a3}. This technique can improve performance in specialized fields such as law, medicine, or science, where precise language and factual accuracy are critical. Additionally, active learning—where the model selectively queries for more data in areas of uncertainty—can further improve data quality and enhance specific capabilities\cite{jeong2024medicaladaptationlargelanguage}.

\subsubsection{Balancing Faithfulness and Flexibility}

While reducing faithfulness-related hallucinations, caution is required to prevent an increase in factual hallucinations caused by poor confidence calibration. Striking a balance between faithfulness—ensuring outputs remain both factually accurate and logically coherent—and flexibility—permitting the model to adeptly manage a wide array of prompts while adapting seamlessly to various contexts—remains a pivotal challenge.A model overly focused on faithfulness may become too conservative, withholding useful insights when faced with ambiguous or uncommon prompts. Conversely, excessive flexibility can lead to overconfident responses that are factually incorrect or logically incoherent.

Hybrid models that combine retrieval-augmented generation (RAG) with standard LLM techniques have shown promise in addressing these issues. RAG enables models to ground responses in external, up-to-date knowledge bases, reducing the risk of factual errors. However, integrating retrieval mechanisms introduces new challenges, such as ensuring that retrieved information is accurately incorporated into the final output, and that the model can reason effectively in contexts where retrieved knowledge is sparse or absent.

Another approach is self-calibration, where the model monitors its own confidence levels across different tasks and adjusts its outputs accordingly. Paired with faithfulness metrics, self-calibration can help models decide when to provide definitive answers and when to express uncertainty, reducing misleading or incorrect outputs.

\subsubsection{Challenges and Future Directions in Inference}

In the inference phase, LLMs excel at tasks requiring abstract reasoning, managing long-distance dependencies, and zero-shot learning, making them highly effective for tasks like translation, summarization, and creative text generation. However, significant challenges remain in areas such as commonsense reasoning, logical operations, mathematical problem-solving, and nuanced emotional interpretation. These limitations often arise from the model’s difficulty in understanding implicit contextual cues, performing multi-step reasoning, and maintaining internal consistency over long dialogues or texts.

Commonsense reasoning is particularly challenging for LLMs because it involves knowledge of everyday experiences, causal relationships, and social norms—information often absent from training data. Developing this capability may require novel approaches, such as integrating structured knowledge bases (e.g., ConceptNet) or training on narrative-driven datasets that emphasize causal sequences and human interactions.

Logical and mathematical reasoning also pose challenges, as these tasks require step-by-step inference and error correction. Current models struggle with maintaining logical coherence over complex reasoning chains or solving multi-step mathematical problems. Techniques like chain-of-thought prompting (CoT) have been developed to break down complex reasoning tasks into smaller steps, but further improvements are needed.

Emotional interpretation and sentiment analysis are additional areas where LLMs often underperform. Understanding emotions requires context sensitivity, and models frequently struggle to interpret subtext, irony, or culturally specific emotional cues. Future models may benefit from affective computing techniques, which focus on recognizing and interpreting human emotions, or multi-modal training (e.g., combining text with audio or visual cues) to better understand and respond to nuanced emotional contexts.

\subsubsection{Knowledge and Capability Boundaries}

Understanding the knowledge and capability boundaries of LLMs is critical for future research. While LLMs can excel in certain tasks, their limitations are often domain-specific and unpredictable. Developing benchmarks that assess a wide range of model capabilities—from factual recall and logical consistency to emotional understanding and commonsense reasoning—is necessary for systematically identifying these boundaries. Additionally, investigating how models recognize their own knowledge gaps—i.e., knowing what they don’t know—could enable better alignment of outputs with user expectations and improve reliability.

Modular LLM architectures, where specialized sub-models are invoked for domain-specific tasks, offer another promising direction. For instance, a general-purpose LLM might defer to a specialized model for tasks requiring legal reasoning or medical diagnostics, ensuring that responses are both accurate and contextually appropriate. This modular approach could help overcome the current limitations of generalized knowledge in LLMs, ensuring more accurate and relevant responses.

In summary, enhancing data quality, balancing faithfulness and flexibility, advancing inference capabilities, and understanding knowledge boundaries are essential for overcoming current LLM limitations. By addressing these challenges, the field can progress towards more accurate, reliable, and versatile AI systems capable of handling complex, context-dependent tasks with greater trustworthiness and insight.

\subsection{Abstract Concepts}

Does a ``truthful subspace'' truly exist within LLMs? Could different types of hallucinations be represented within a unified ``hallucination space''?

Understanding the relationship between hallucinations and the space of truth is a complex challenge. Hallucinations are often grounded in reality but deviate from it in critical ways. What geometric or structural connections exist between these two spaces, and what practical applications could these insights offer? Beyond the binary of truth and hallucination, could there be regions in the space that represent ``super assertions''—statements that defy simple categorization as true or false—or even ``absolute errors'' that are the complete opposite of absolute truths? These unexplored areas in the activation space offer promising directions for deeper investigation.

One potential research avenue is exploring how these conceptual spaces interact within the activation dynamics of LLMs. If we could map the activation space into distinct regions—such as ``truthful,'' ``hallucinatory,'' and ``ambiguous''—it might enable a more refined approach to understanding and controlling LLM behavior. This mapping could not only identify the ``truthful subspace'' but also help develop mechanisms to guide the model toward this space during inference, reducing the likelihood of hallucinations.

The relationship between the spaces of hallucination and truth can be regarded as a complex geometric problem. In the activation space, hallucinations and realities may interweave, forming an intricate network.
A significant challenge is clearly defining and demarcating these subspaces.
The borders between truth and hallucination may be blurred, as models often produce responses that are partially true or contextually dependent.
For example, statements that are factual in one domain or culture may be seen as false in another. Developing a geometric or algebraic framework for analyzing these regions could provide valuable insights into how different types of hallucinations arise.
It might also reveal whether all hallucinations share a common structural pattern within the LLM or if distinct types, such as factual errors and logical inconsistencies, occupy different subspaces.

The hypothesis of a ``super assertion'' space—statements that resist easy classification—opens further questions about undecidable outputs in LLMs. Such statements, which may not strictly align with truth or hallucination, could represent high-entropy content that the models struggle to resolve within their training paradigms. The existence of this space challenges the binary notion of ``truth versus hallucination'' and suggests a spectrum of certainty or confidence within the model’s output. Understanding how LLMs handle these ambiguous statements could be crucial for calibrating their responses, especially in domains that require high precision or interpretative nuance.

Similarly, the concept of ``absolute errors''—utterances that are fundamentally incorrect or nonsensical in any context—suggests that part of the activation space may be dedicated to outputs that fail to align with any coherent structure of knowledge or logic. Investigating the nature of these absolute errors, and how they are represented within LLM activations, could inform the development of filters or constraints to suppress them during generation.

Another intriguing research direction involves exploring the potential pathways between these subspaces during model training and inference. For example, does a model’s traversal through its activation space reflect a gradual transition from a ``truthful'' region to a ``hallucinatory'' one as its confidence in a response decreases or as it encounters more ambiguous data? If so, could this transition be detected and controlled to prevent the model from drifting into less reliable territories? Understanding these pathways and their triggers could be key to developing more robust LLMs that not only recognize their own knowledge boundaries but actively avoid exceeding them.

In addition, the relationship between ``truthful subspaces'' and model architecture warrants further exploration. How do architectural modifications—such as changes in attention mechanisms, layer composition, or fine-tuning strategies—impact the size, shape, and accessibility of these subspaces? This line of inquiry could reveal design principles for constructing LLMs that inherently favor truthful and coherent outputs, thus minimizing hallucinations by design rather than through post-processing techniques.

Finally, examining how these subspaces evolve across different LLMs of varying sizes, training datasets, and domains may uncover universal patterns or principles that govern hallucination phenomena. This comparative approach could also shed light on whether certain architectural or training choices predispose models to specific types of hallucinations, guiding future model development toward more reliable and controllable behaviors.

In summary, the exploration of abstract concepts such as ``truthful subspaces,'' ``hallucination spaces,'' and ``super assertions'' presents an exciting frontier in the study of LLMs. Understanding and defining these spaces could lead to more effective methods for managing hallucinations and improving model reliability.

\subsection{Method Exploration}

Current approaches to mitigating hallucinations each have their limitations. For example, Retrieval-Augmented Generation (RAG) poses risks to the retrieval ecosystem. When RAG-generated content is added to the knowledge base, it can be repeatedly retrieved and reused. If this content is flawed, hallucinations may perpetuate without correction, disrupting the self-regulating dynamics of the retrieval ecosystem and hindering its sustainable growth \cite{chen2024spiral}.

\subsubsection{Challenges in Fine-Tuning and Subspace Exploration}

Fine-tuning (FT) is widely adopted to enhance LLM performance for specific domains or tasks. However, it has not fully addressed the concept of the ``truthful subspace'' within the activation space. A key challenge is identifying whether such a subspace exists, and if so, how to navigate and leverage it effectively. This subspace could represent regions where the model's internal representations align with factual, reliable content. However, identifying it is non-trivial, requiring differentiation between accurate interpretations of truth and those that produce hallucinations or logical inconsistencies.

An interesting avenue for exploration is the potential existence of a unified ``hallucination space'' within the model's activation dynamics. This space might encompass various hallucinations, from factual inaccuracies to logical fallacies, raising questions about how these different errors are represented and related. If structural commonalities exist between hallucinations, it could lead to more efficient mitigation strategies that address multiple error types simultaneously.

The presence of ``super assertions''—statements that challenge binary truth classifications—complicates this exploration. These statements, often based on abstract or context-dependent interpretations, may lie on the boundary between truth and hallucination. Developing methods to handle these borderline cases is essential for improving LLM robustness. Similarly, identifying ``absolute errors,'' which fundamentally oppose truth, can help define the boundaries of what LLMs should avoid generating. Mapping these abstract spaces within the activation dynamics could lead to better error correction and model reliability.

\subsubsection{Addressing Confidence Calibration, Prompt Engineering, and Decoding Strategies}

While confidence calibration and FT aim to align LLM responses with factual accuracy and self-awareness, they also risk altering the model’s core capabilities. Over-calibration can lead to overly cautious models that withhold valid answers, while under-calibration may result in unjustified confidence in incorrect responses. It is crucial to balance calibration with the model's ability to explore diverse outputs and handle ambiguous inputs, ensuring the model remains flexible across various contexts and tasks without sacrificing accuracy.

Prompt engineering, which steers LLMs toward better responses, has been successful but faces stability and reliability issues. Prompts are sensitive to wording, structure, and context, leading to inconsistent outputs with minor input changes. Developing standardized prompt construction methodologies that ensure robustness and consistent results across diverse scenarios is an active research area. Additionally, understanding how prompt templates interact with LLMs at a structural level could reduce hallucinations and improve coherence in responses.

Decoding strategies, such as beam search, top-k sampling, and temperature control, aim to enhance response quality by guiding the generation process. However, these strategies often lack cross-task generalizability, meaning a method effective for one task may underperform on another. This limitation is particularly evident in complex reasoning tasks where logical consistency is critical. Future decoding strategies may involve dynamic methods that adapt to each task’s requirements, potentially incorporating feedback loops or real-time calibration based on model confidence and external verification.

\subsubsection{Towards Novel Solutions: Integrative and Multimodal Approaches}

The limitations of current methods highlight the need for integrative approaches to mitigate hallucinations. One direction is using hybrid techniques that combine methods like FT, confidence calibration, prompt engineering, and retrieval augmentation into a unified framework. By leveraging each method’s strengths and compensating for their weaknesses, a more robust system could emerge, better equipped to handle hallucinations across diverse applications.

Exploring multimodal data—such as visual, audio, or contextual cues—in training and inference processes may also help reduce hallucinations. For instance, grounding language generation in visual information could improve the model’s contextual understanding, leading to more accurate outputs. This approach would be especially useful for tasks requiring an understanding of real-world events or complex narrative sequences.

Another promising development is self-reflective models, which critique their outputs, identify inconsistencies, and adjust responses accordingly. This feedback loop could reduce hallucinations, especially when combined with external verification mechanisms. Such self-improving capabilities allow models to learn not only from their training data but also from real-world interactions, enhancing accuracy and reliability over time.

\subsubsection{Open Questions and Future Directions}

The complex relationship between truth and hallucination in LLMs presents several research opportunities. One key area is understanding how truthful and hallucination subspaces evolve during training and how different objectives influence them. Additionally, exploring how architectural modifications, such as attention mechanisms or layer adjustments, affect a model's ability to navigate these subspaces could lead to more reliable models.

Another important direction is improving models' ability to assess and express their uncertainty. A deeper understanding of confidence calibration and how it relates to generation could lead to more transparent AI systems where models indicate when they are unsure or need verification.

Finally, interdisciplinary approaches combining insights from linguistics, cognitive science, and human-computer interaction could deepen our understanding of hallucinations. By studying how humans generate and detect communication errors, novel techniques for improving LLMs may emerge, reducing their tendency to hallucinate. Ultimately, combining methodological improvements, abstract conceptual explorations, and cross-disciplinary insights will be crucial for advancing LLMs towards more accurate and trustworthy language generation.

\section{Conclusion}

This survey provides an in-depth exploration of the prevalent phenomenon of ``hallucinations'' in large language models (LLMs). It thoroughly defines the concept, breaking down different types of hallucinations and distinguishing their various manifestations to better understand their nature. The internal mechanisms of LLMs are closely analyzed to uncover the root causes of hallucinations, using mathematical methods to interpret the complex principles underlying their emergence and development.

Regarding detection and evaluation, this survey presents a range of efficient methods and metrics tailored to different types of hallucinations. These tools enable researchers and practitioners to conduct quantitative analyses, assessing the extent and impact of hallucinations across various contexts. With a clear understanding of hallucination causes, the survey explores multiple mitigation strategies, addressing both structural and functional aspects of LLMs to improve their reliability, performance, and practical applications.

Looking ahead, the survey highlights key research gaps and challenges in the field, pointing out areas where current methodologies fall short and where further investigation is needed. It outlines potential research directions, aiming to inspire further exploration and technological innovation within the academic community. Ultimately, this work seeks to establish a comprehensive research framework on LLM hallucinations, offering both theoretical insights and practical resources for practitioners to effectively address these complex challenges.
\ifCLASSOPTIONcaptionsoff
  \newpage
\fi



\bibliographystyle{IEEEtran}
\bibliography{Loki}
%

%




\end{document}